\documentclass[10pt,twocolumn,letterpaper]{article}

\usepackage[pagenumbers]{cvpr} %

\usepackage[dvipsnames]{xcolor}

\usepackage{times}
\usepackage{epsfig}
\usepackage{graphicx}
\usepackage{amsmath}
\usepackage{amssymb}
\usepackage{picture}
\usepackage{booktabs}
\usepackage{caption}
\usepackage{fontawesome}
\usepackage{multirow}
\usepackage{subcaption}
\usepackage{bm}
\usepackage{bbm}
\usepackage{tikz}
\usepackage{xcolor}
\usepackage{wrapfig}

\definecolor{cvprblue}{rgb}{0.21,0.49,0.74}
\usepackage[pagebackref,breaklinks,colorlinks,citecolor=cvprblue]{hyperref}

\renewcommand*{\ie}{i.e.,\@\xspace}
\renewcommand*{\eg}{e.g.,\@\xspace}

\newcommand*{\x}{\mathsf{x}\mskip1mu}

\newcommand{\stemp}{\faCube\@\xspace}
\newcommand{\skey}{\faPaw\@\xspace}
\newcommand{\sview}{\faCameraRetro\@\xspace}
\newcommand{\smask}{\faScissors\@\xspace}
\newcommand{\svid}{\faVideoCamera\@\xspace}
\newcommand{\sflow}{\faForward\@\xspace}
\newcommand{\sdepth}{\faEraser\@\xspace}
\newcommand{\sssl}{\faCompress\@\xspace}
\newcommand{\skel}{\faMale\@\xspace}

\newcommand{\newstar}{\hspace{-2pt}$^*$}

\def\confName{CVPR}
\def\confYear{2024}

\begin{document}

\title{SAOR: Single-View Articulated Object Reconstruction}

\author{Mehmet Ayg\"un
\quad Oisin Mac Aodha \\ [0.3em]
University of Edinburgh\\ [0.1em]
\small\url{mehmetaygun.github.io/saor}
}

\twocolumn[\maketitle\vspace{-2em}\begin{center}

    \begin{picture}(0.49\textwidth, 5)
    \put(0.03\textwidth,0){Input}
    \put(0.12\textwidth,0){Pose}
    \put(0.24\textwidth,0){Reconstruction}
    \put(0.41\textwidth,0){Parts}
    \end{picture}
    \begin{picture}(0.49\textwidth, 5)
    \put(0.035\textwidth,0){Input}
    \put(0.13\textwidth,0){Pose}
    \put(0.24\textwidth,0){Reconstruction}
    \put(0.41\textwidth,0){Parts}
    \end{picture}
    \vspace{-5pt}
    \includegraphics[width=0.4975\textwidth, trim=5 0 5 0, clip]{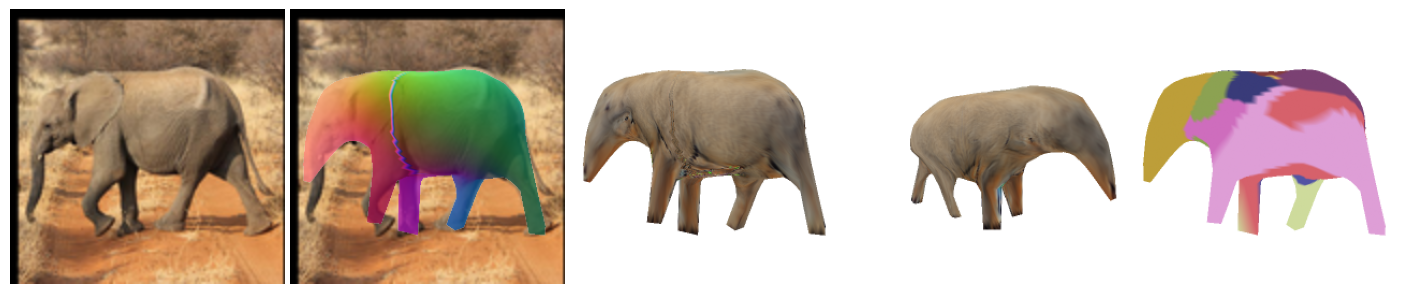}
    \includegraphics[width=0.4975\textwidth, trim=5 0 5 0, clip]{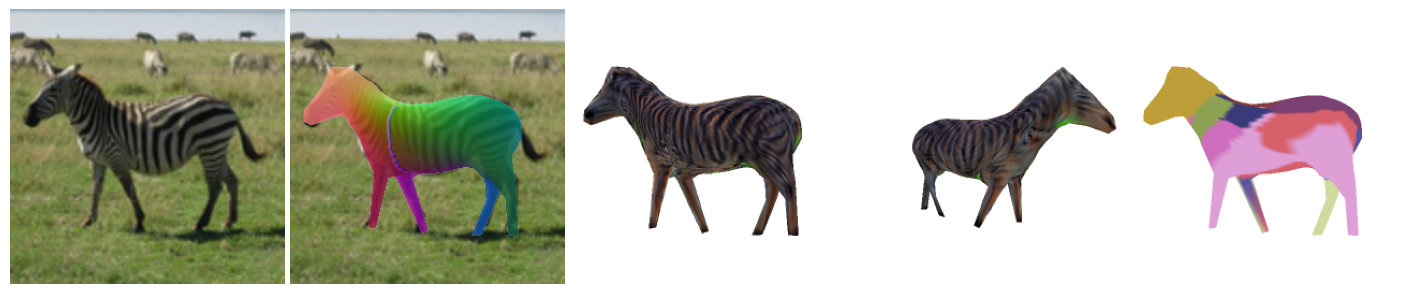}
    \vspace{-5pt}
   \includegraphics[width=0.4975\textwidth, trim=5 0 5 0, clip]{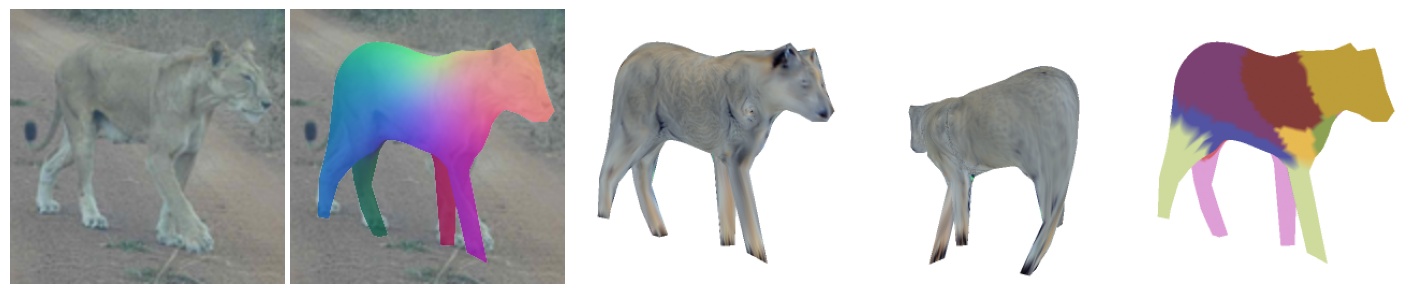}
    \includegraphics[width=0.4975\textwidth, trim=5 0 5 0, clip]{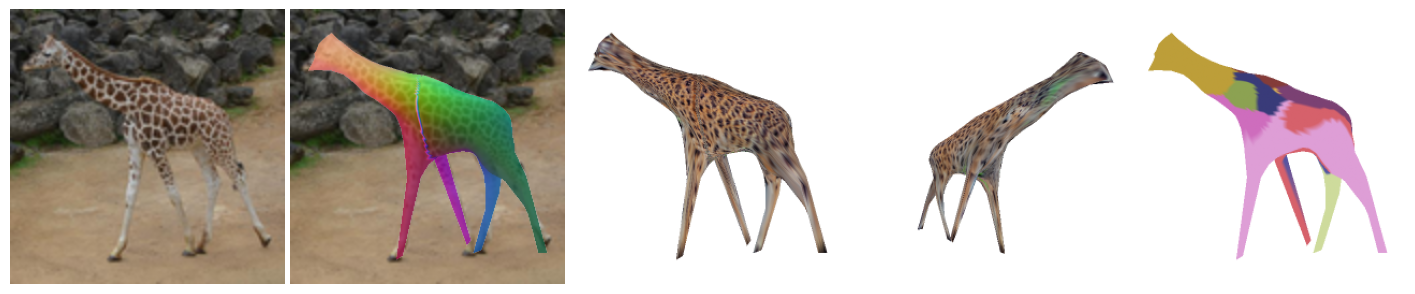}
   \vspace{-5pt}
    \includegraphics[width=0.4975\textwidth, trim=5 0 5 0, clip]{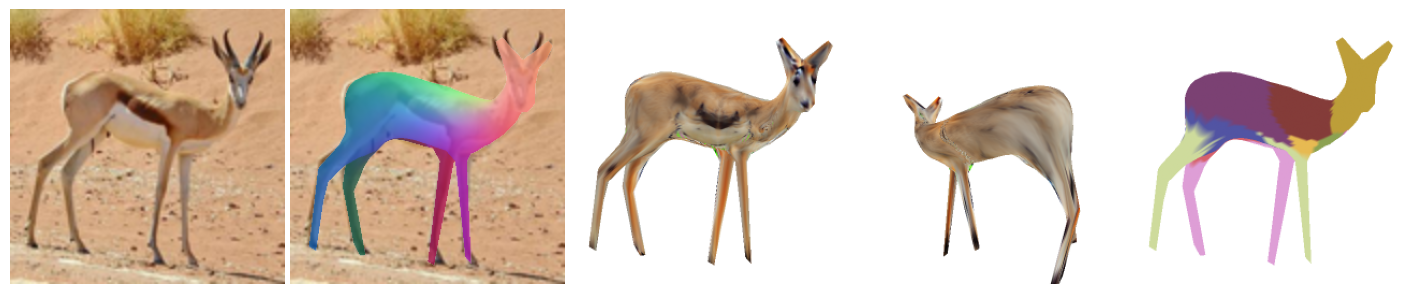}
    \includegraphics[width=0.4975\textwidth, trim=5 0 5 0, clip]{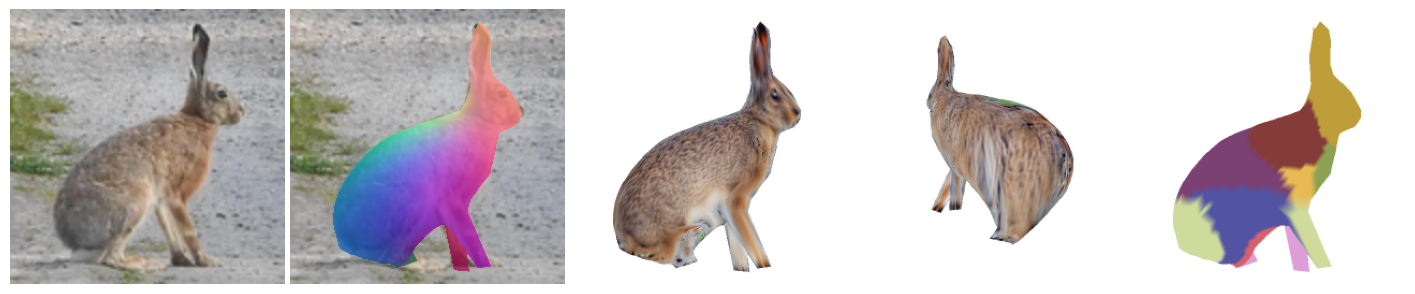}
    \vspace{-5pt}
    \includegraphics[width=0.4975\textwidth, trim=5 0 5 0, clip]{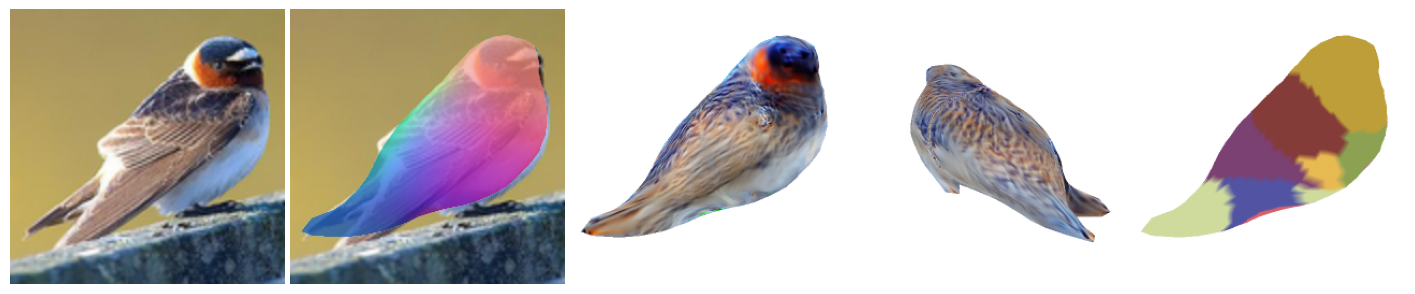}
    \includegraphics[width=0.4975\textwidth, trim=5 0 5 0, clip]{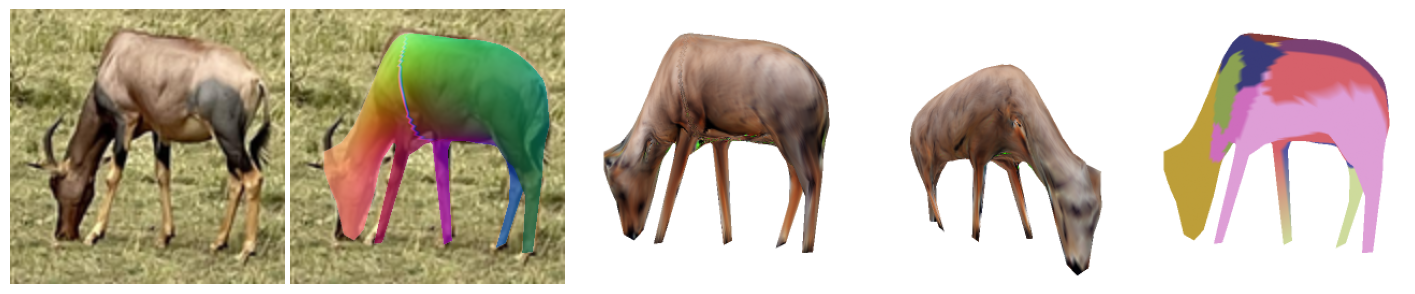}
    \vspace{-5pt}

\end{center}
\vspace{-20pt}
\captionof{figure}{SAOR capable of predicting the 3D shape of an articulated object category from a single image. 
Our model is trained on multiple categories simultaneously using self-supervision on single-view image collections. It can efficiently predict object pose, 3D shape reconstruction, and unsupervised part-level assignment using only a single forward pass per image at test time in a category-agnostic way.
}
\label{fig:teaser}
\bigbreak]

\begin{abstract}
We introduce SAOR, a novel approach for estimating the 3D shape, texture, and viewpoint of an articulated object from a single image captured in the wild. 
Unlike prior approaches that rely on pre-defined category-specific 3D templates or tailored 3D skeletons, SAOR learns to articulate shapes from single-view image collections with a skeleton-free part-based model without requiring any 3D object shape priors. 
To prevent ill-posed solutions, we propose a cross-instance consistency loss that exploits disentangled object shape deformation and articulation. 
This is helped by a new silhouette-based sampling mechanism to enhance viewpoint diversity during training. 
Our method only requires estimated object silhouettes and relative depth maps from off-the-shelf pre-trained networks during training. 
At inference time, given a single-view image, it efficiently outputs an explicit mesh representation. 
We obtain improved qualitative and quantitative results on challenging quadruped animals compared to relevant existing work. 
\end{abstract}

\vspace{-8pt}
\section{Introduction}
\label{sec:intro}
\vspace{-5pt}
Considered as one of the first PhD theses in computer vision, Roberts~\cite{roberts1963machine} aimed to reconstruct 3D objects from single-view images. 
Despite significant progress in the preceding sixty years \cite{blanz1999morphable,cashman2012shape,kar2015category,kanazawa2018learning}, the problem remains very challenging, especially for highly deformable categories photographed in the wild, \eg animals. 
In contrast, humans can infer the 3D shape of an object from a single image by making use of priors about the natural world and familiarity with the object category present. 
Some of these natural-world low-level priors can be explicitly defined (\eg symmetry or smoothness), but manually encoding and utilizing high-level priors (\eg 3D category shape templates) for all categories of interest is not a straightforward task. 

Recently, multiple methods have attempted to learn 3D shape by making use of advances in deep learning and progress in differentiable rendering~\cite{loper2014opendr, kato2018neural, liu2019soft}.
This has resulted in impressive results for synthetic man-made categories~\cite{choy20163d,kato2018neural,wang2018pixel2mesh} and humans~\cite{loper2015smpl, guler2018densepose}, where full or partial 3D supervision is readily available. 
However, when 3D supervision is not available, the reconstruction of articulated object classes remains challenging. 
This is due to factors such as: (i) methods not modeling articulation~\cite{kanazawa2018learning,kulkarni2019csm,goel2020shape,monnier2022share}, (ii) the reliance on category-specific 3D template~\cite{kokkinos2021learning,kulkarni2020acsm,zuffi2019three} or manually defined 3D skeleton supervision~\cite{wu2021dove,wu2022magicpony}, or (iii) requiring multi-view training data such as video~\cite{wu2021dove, kokkinos2021learning, yang2021lasr}. 

In this paper, we introduce \textbf{SAOR}, a novel self-supervised \textbf{S}ingle-view \textbf{A}rticulated \textbf{O}bject \textbf{R}econstruction method that can estimate the 3D shape of articulating object categories, \eg animals. 
We forgo the need for explicit 3D object shape or skeleton supervision at training time by making use of the following assumption: \textit{objects are made of parts, and these parts move together}. 
Given a single input image, our proposed method predicts the 3D shape of the object and partitions it into parts. 
It also predicts the transformation for each part and deforms the initially estimated shape, in a skeleton-free manner, using a linear skinning approach. 
We only require easy to obtain information derived from single-view images during training, \eg estimated object silhouettes and predicted relative depth maps. 
SAOR is trained end-to-end, and outputs articulated 3D object shape, texture, 3D part assignment, and camera viewpoint. 
Example qualitative results can be seen in Fig.~\ref{fig:teaser}.

We make the following contributions: (i) We demonstrate that articulation can be learned using image-based self-supervision alone via our new part-based SAOR approach which is trained on multiple categories simultaneously without requiring any 3D template or skeleton prior. 
(ii) As estimating the 3D shape of an articulated object from a single image is an under-constrained problem, we introduce a cross-instance swap consistency loss that leverages our disentanglement of shape deformation and articulation, in addition to a new silhouette-based sampling mechanism, that enhances the diversity of object viewpoints sampled during training.
(iii) We illustrate the effectiveness of our approach on a diverse set of over 100 challenging categories covering quadrupeds and bipeds, and present quantitative results where we outperform existing methods that do not use explicit 3D supervision. 
Code will be made available.

\section{Related Work}
\label{sec:rel_work}
\vspace{-5pt}
Here we discuss works that attempt to estimate the 3D shape of an object in a single image using image-based 2D supervision during training. 
We do not focus on works that require explicit 3D supervision~\cite{choy20163d,kato2018neural,wang2018pixel2mesh,mescheder2019occupancy} or multi-view images for training~\cite{yu2021pixelnerf,jain2021putting,vasudev2022ss3d,liu2023zero}. We also do not cover methods that only reconstruct single object instances \cite{mildenhall2021nerf,park2021nerfies,poole2022dreamfusion} or models for multi-object scenes~\cite{niemeyer2021giraffe}. 
For a recent overview of related topics, we refer readers to~\cite{tretschk2022state}.

\noindent\textbf{Deformable 3D Models.} The pioneering work of Blanz and Vetter~\cite{blanz1999morphable} marked the introduction of deformable models to represent the 3D shape of an object category using vector spaces. 
By using 3D scans of human faces, they created a deformable model which captured inter-subject shape variation and demonstrated the ability to reconstruct 3D faces from unseen single-view images. 
This concept was later expanded to more complex shapes such as the human body~\cite{loper2015smpl,anguelov2005scape}, hands~\cite{taylor2014user,khamis2015learning}, and animals~\cite{zuffi20173d}. 

Recent work has combined deep learning with 3D deformable models \cite{loper2015smpl,zuffi2019three,biggs2020left,rueegg2022barc} to predict the shape of articulated objects from single-view input images. 
Given an input image, these methods estimate the parameters of a known deformable 3D model and render the object using the predicted camera viewpoint. 
Although this line of work has led to impressive results for the human body~\cite{loper2015smpl}, the results for deformable animal categories are lacking \cite{zuffi2019three,biggs2020left,rueegg2022barc}. 
This is because popular human deformable models, \eg SMPL~\cite{loper2015smpl}, are constructed using thousands of high-quality real human 3D scans. 
In contrast, animal focused 3D models, \eg SMAL~\cite{zuffi2019three}, are generated using 3D scans from a small number of toy animals. 

The above models are parameter-efficient due to their low dimensional shape parameterization, which facilitates easier optimization. 
However, beyond common categories, such as dogs~\cite{rueegg2022barc}, it can be prohibitively difficult to find 3D scans for each new object category of interest.
In this work, we eliminate the need for prior 3D scans of objects by combining linear vertex deformation with a skeleton-free~\cite{liao2022skeleton} linear blend skinning~\cite{lewis2000pose} approach to model the 3D shape of articulated objects using only images at training time.

\noindent \textbf{Unsupervised Learning of 3D Shape.} 
To overcome the need for large collections of aligned 3D scans from an object category of interest, there has been a growing body of work that attempts to learn 3D shape using images from only minimal, if any, 3D supervision. 
The common theme of these methods is that they treat shape estimation as an image synthesis task during training while enforcing geometric constraints on the rendering process. 

One of the first object-centric deep learning-based methods to not use dense 3D shape supervision for single-view reconstruction was CMR~\cite{kanazawa2018learning}. 
CMR utilizes camera pose supervision estimated from structure from motion, along with human-provided 2D semantic keypoint supervision during training and a coarse template mesh initialized from the keypoints. 
Subsequently, U-CMR~\cite{goel2020shape} remove the keypoint supervision by using a multi-camera hypothesis approach which assigns and optimizes multiple cameras for each instance during training. 
IMR~\cite{tulsiani2020implicit} starts from a category-level 3D template and learns to estimate shape and camera viewpoint from images and segmentation masks.
UMR~\cite{li2020self} enforces consistency between per-instance unsupervised 2D part segmentations and 3D shape. 
They do not assume access to a 3D shape template (or keypoints) but instead learn one via iterative training. 
SMR~\cite{hu2021self} also uses object part segmentation from a self-supervised network as weak supervision. 
Shelf-SS~\cite{ye2021shelf} uses a semi-implicit volumetric representation and obtains consistent multi-view reconstructions using generative models similar to~\cite{henzler2019escaping}.
Like us, all of these methods use object silhouettes (\ie foreground masks) as supervision.

Recently, Unicorn~\cite{monnier2022share} combined curriculum learning with a cross-instance swap loss to help encourage approximate multi-view consistency across object instances when training a reconstruction network without silhouettes. 
Their swap loss makes use of an online memory bank to select pairs of images that contain similar shape or texture. 
The pairs are restricted to be observed from different estimated viewpoints. 
Then a consistency loss is applied which explicitly forces pairs to share the same shape or texture. 
In essence, this is a form of weak multi-view supervision under the assumption that the shape of the object pair are the same. 
However, this assumption breaks down for articulating objects. 
Inspired by this, we propose a more efficient and effective swap loss designed for articulating objects. 

There are also approaches that predict a mapping from image pixels to the surface of a 3D object template as in~\cite{guler2018densepose,neverova20continuous}. 
CSM~\cite{kulkarni2019csm} eliminates the need for large-scale 2D to 3D surface annotations via an unsupervised 2D to 3D cycle consistency loss. 
The goal of their loss is to minimize the discrepancy between a pixel location and a corresponding 3D surface point that is reprojection based on the estimated camera viewpoint. 
In contrast, we do not require any 3D templates or manually defined 2D annotations. 

\noindent\textbf{Learning Articulated 3D Shape.} 
Most natural object categories are non-rigid and can thus exhibit some form of articulation. 
This natural shape variation between individual object instances violates the simplifying assumptions made by approaches that do not attempt to model articulation. 

A-CSM~\cite{kulkarni2020acsm} extends CSM~\cite{kulkarni2019csm} by making the learned mapping articulation aware. 
Given a 3D template of the object category, they first manually define the parts of the object category and a hierarchy between the parts. 
Then, given an input image, they predict transformation parameters for each part so they can articulate the initial 3D template before calculating the mapping between the 3D template and the input pixels. 
Recently \cite{stathopoulos2023learning} show that A-CSM can be trained with noisy keypoint labels. Instead of manually defining parts, \cite{kokkinos2021learning} initialize sparse handling points, predict displacements for these points, and articulate the shape using differentiable Laplacian deformation. 
However, each of these methods requires a pre-defined 3D template of the object category. 

DOVE~\cite{wu2021dove}, LASSIE~\cite{yao2022lassie}, and MagicPony~\cite{wu2022magicpony} are recent methods that are capable of predicting the 3D geometry of articulated objects without requiring a 3D category template shape. 
However, they require a predefined category-level 3D skeleton prior in order to model articulating object parts such as legs. 
While 3D skeletons are easier to define compared to full 3D shapes, they still need to be provided for each object category of interest and have to be tailored to the specifics of each category, \eg the trunk of the elephant is not present in other quadrupeds. 
In the case of MagicPony~\cite{wu2022magicpony}, in addition to the skeleton and its connectivity, per-bone articulation constraints are also provided, which necessitates more manual labor. 
Additionally, a single skeleton may be insufficient if there are large shape changes exhibited across instances of the category. 

MagicPony~\cite{wu2022magicpony} builds on DOVE~\cite{wu2021dove}, by removing the need for explicit video data during training. 
Inspired by UMR~\cite{li2020self}, MagicPony makes use of weak correspondence supervision from a pre-trained self-supervised network to enforce pixel-level consistency between 2D images and learned 3D shape. Concurrent to our work 3D-Fauna~\cite{li2024learning} extends MagicPony for quadrupeds in a multi-category setting. LASSIE~\cite{yao2022lassie} is another skeleton-based approach that uses correspondence information from self-supervised features and manually pre-defined part primitives. 
Like us, they model object parts, but their goal is not to learn a model that can directly predict shape from a single image. 
Instead, their approach learns instance shape from a set of images via test-time optimization. 
In recent work, \cite{yao2022hi-lassie} automatically extracts the skeleton from a user-defined canonical image, but still requires test-time optimization. 

We train with single-view image collections, but there are also several works that use video as a data source for modeling articulating objects~\cite{li2020online,wu2021dove,yang2021lasr,yang2021viser} and other methods that perform expensive test-time optimization for fitting or refinement~\cite{zuffi2019three,kokkinos2021point,li2020online,wu2022magicpony,yao2022lassie}. 
In contrast, we only require self-supervision derived from single-view images and our inference step is performed efficiently via a single forward pass through a deep network.

\vspace{-5pt}
\section{Method}
\label{sec:method}
\vspace{-5pt}

\begin{figure*}[t]
  \centering
  \includegraphics[width=\textwidth]{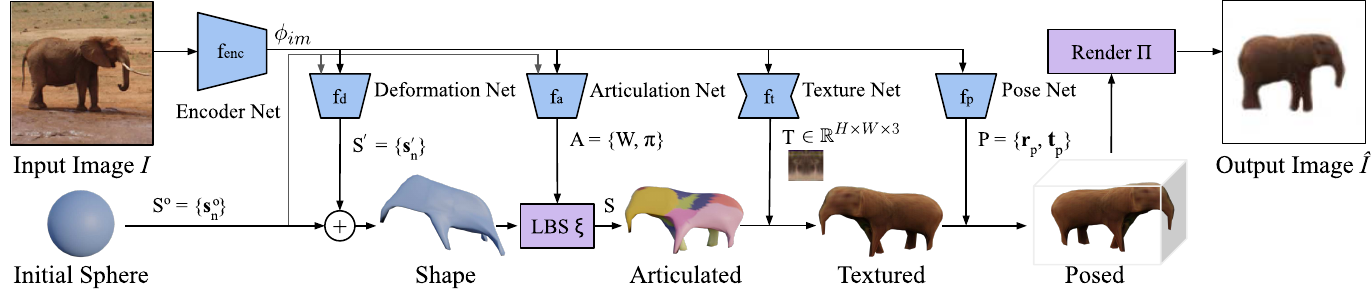}
  \vspace{-15pt}
  \caption{Overview of the generation phase of our SAOR method. 
  Given a single image $I$ as input, we extract a global feature vector~$\phi_{im}$ which is decoded by four separate networks ($f_d$, $f_a$, $f_t$, and $f_p$) to generate a final output image $\hat{I}$. 
  We start by deforming an initial sphere, articulate it using a part-based linear blend skinning (LBS) operation $\xi$, texture the mesh, and render it using a differential render $\Pi$ so that it is depicted from the same viewpoint as the input image. 
  The parameters for each of the networks presented are trained in an end-to-end manner using image reconstruction-based self-supervision from multiple different categories using the same model.}
  \label{fig:main_method}
  \vspace{-12pt}
\end{figure*}

Our objective is to estimate the shape $S$, texture $T$, and camera pose (\ie viewpoint) $P$ of an object from an input image $I$. To accomplish this, we employ a self-supervised analysis-by-synthesis framework \cite{grenander78,kulkarni2015deep} which reconstructs images using a differentiable rendering operation, denoted as $\hat{I}=\Pi(S, T, P)$. The model is optimized by minimizing the discrepancy between a real image $I$ and the corresponding rendered one $\hat{I}$. 
In this section, we describe how the above quantities are estimated to ensure that the predicted 3D shape is plausible. 
An overview of the generation phase of our method can be seen in Fig.~\ref{fig:main_method}

\subsection{SAOR Model}
\vspace{-5pt}
Taking inspiration from previous works \cite{kanazawa2018learning,ye2021shelf,monnier2022share}, we initialize a sphere-shaped mesh with initial vertices $S^{\circ}$ with fixed connectivity.
We then extract a global image representation $\phi_{im} = f_{enc} (I) \in{\mathbb{R}^D}$ using a neural network encoding function. 
From this, we utilize several modules, described below, to predict the shape deformation, articulation, camera viewpoint, and object texture necessary to generate the final target shape.

\noindent\textbf{Shape.} We predict the object shape by deforming and articulating an initial sphere mesh $S^{\circ} = \{\bm{s}^\circ_{n}\}_n^N$. 
Here, each of the $N$ elements of $S^{\circ}$ are 3D coordinates. 
We estimate the vertices of the deformed shape using a deformation function $\bm{s}'_{i} = \bm{s}_i^{\circ} + f_{d}(\bm{s}^{\circ}_{i}, \phi_{im})$, which outputs the displacement vector for the initial points. 
The deformation function $f_{d}$ is modeled as a functional field, which is a 3-layer MLP similar to~\cite{niemeyer2021giraffe,monnier2022share}. 
As most natural objects exhibit bilateral symmetry, similar to~\cite{kanazawa2018learning}, we only deform the vertices of the zero-centered initial shape that are located on the positive side of the xy-plane and reflect the deformation for the vertices on the negative side.
We then articulate the deformed shape using linear skinning~\cite{lewis2000pose} in a skeleton-free manner~\cite{liao2022skeleton} to obtain the final shape $S = \xi (S', A)$, where $A$ is the output of our articulation prediction function, which we describe in more detail later in Sec. \ref{sec:skeleton}.

\noindent \textbf{Texture.} To predict the texture of the object, we generate a UV image by transforming the global image feature, $T = f_{t}(\phi_{im})$. 
The function $f_{t}$ is implemented as a convolutional decoder, which maps a one-dimensional input representation to a texture map, $f_{t}: \mathbb{R}^{D} \mapsto \mathbb{R}^{H \times W \times 3}$. 
This approach is similar to previous works \cite{monnier2022share,niemeyer2021giraffe}. 
However, unlike existing work~\cite{kanazawa2018learning,li2020self} that copy the pixel colors of the input image directly to create a texture image using a predicted flow field, we predict texture directly. 
In initial experiments, we found that estimating texture flow only gave minimal improvements, for an increase in complexity. 

\noindent \textbf{Camera Pose.} We use Euler angles (azimuth, elevation, and roll) along with camera translation to predict the camera pose, similar to previous works \cite{goel2020shape, monnier2022share}. Instead of using multiple camera hypotheses for each input instance~\cite{monnier2022share}, for each forward pass, or optimizing them for each training instance~\cite{goel2020shape}, we use several camera pose predictors, but only select the one with the highest confidence score for each forward pass, as described in~\cite{wu2022magicpony}. 
Specifically, we predict the camera pose as $P \in \mathbb{R}^{6} = f_{p}(\phi_{im})$. 
Here, $P = {\bm{r}_p, \bm{t}_p}$ represents the predicted camera rotation and translation. 
This approach accelerates the training process and reduces memory requirements since we only need to compute the loss for one camera in each forward pass. 
We only incorporate priors about the ranges of elevation and roll predictions, instead of a strong uniformity constraint on the distribution of the camera poses as in~\cite{monnier2022share} or fixed elevation as in~\cite{wu2022magicpony}. 

\subsection{Skeleton-Free Articulation}
\label{sec:skeleton}
Many natural world object categories exhibit some form of articulation, \eg the legs of an animal. 
Existing work has attempted to model this via deformable 3D template models~\cite{rueegg2022barc} or by using manually defined category-level skeleton priors~\cite{wu2021dove,wu2022magicpony}. 
However, this assumes one has access to category-level 3D supervision during training. 
This would be difficult to obtain in our setting as we train on over 100 categories simultaneously. 
We instead propose a skeleton-free approach by modeling articulation using a part-based model. 
Our approach is inspired by~\cite{liao2022skeleton}, who proposed a related skeleton-free representation for the task of pose transfer between 3D meshes. 
However, in our case, we train a model that can predict parts in an image from self-supervision alone. 

Our core idea is to partition the 3D shape into parts and deform each part based on predicted transformations. 
To achieve this, we predict a part assignment matrix $W \in \mathbb{R}^{ N \times K}$, that represents how likely it is that a vertex belongs to a particular part, where $\sum^{K}_{k} W_{i,k}= 1$. 
Here, $K$ is a hyperparameter that represents the number of parts and $N$ is the number of vertices in the mesh. We also predict transformation parameters $\bm{\pi} = \{(\bm{z}_k,\bm{r}_k,\bm{t}_k)\}_{k}^K$ for each part which consists of scale $\bm{z}_k \in \mathbb{R}^{3}$,
rotation $\bm{r}_k \in \mathbb{R}^{3 \times 3}$, and translation $\bm{t}_k \in \mathbb{R}^{3}$.
Each of these parameters are predicted using different MLPs that take the global image feature $\phi_{im}$ as input and output $f_{a}(S^\circ, \phi_{im}) = A = \{ W, \bm{\pi}\}$. 

Articulation can be applied to a shape using a set of deformations using the linear blend skinning equation~\cite{jacobson2014skinning}. 
Here, each vertex needs to be associated with deformations by the skinning weights. 
In previous work~\cite{wu2021dove,yao2022lassie,wu2022magicpony}, skinning weights are calculated using a skeleton prior (\eg a set of bones and their connectivity). 
We instead estimate skinning weights using a part-based model that does not require a prior skeleton or any ground truth part segmentations. 
We first calculate the centers for each part from the vertices of the deformed shape $\bm{s}'_{i} \in S'$,

\begin{equation}
\bm{c}_{k} = \frac{\sum^{N}_{i} \bm{s}'_{i} * W_{i,k}}{\sum^{N}_{i} W_{i,k}}. 
\end{equation} The final position of a vertex $\bm{s_{i}}$ for the final shape $S$ is then calculated using the skinning weight of the vertex and estimated part transformations as

\begin{equation}
  \bm{s}_{i} = \sum^{K}_{k} W_{i,k} \bm{z}_{k} \odot (\bm{r}_{k} (\bm{s}'_{i} - \bm{c}_{k}) + \bm{t}_{k}),  
\end{equation} where $\bm{z}_{k}$, $\bm{r}_{k}$, and $\bm{t}_{k}$ are the predicted scale, rotation, and translation parameters corresponding to part $k$ and $\odot$ is an element-wise multiplication.
In addition to the reconstruction losses, we apply regularization on the part assignment matrix $W$ that encourages the size of each part segment to be similar for each instance. 
As each of the above operations are differentiable, articulation is learned via self-supervised without requiring any 3D template shapes~\cite{kulkarni2020acsm}, predefined skeletons~\cite{wu2022magicpony}, or part segmentations~\cite{li2020self}.

\subsection{Swap Loss and Balanced Sampling}
\label{sec:swap}

\begin{figure}[t]
\centering
  \includegraphics[width=1.0\columnwidth]{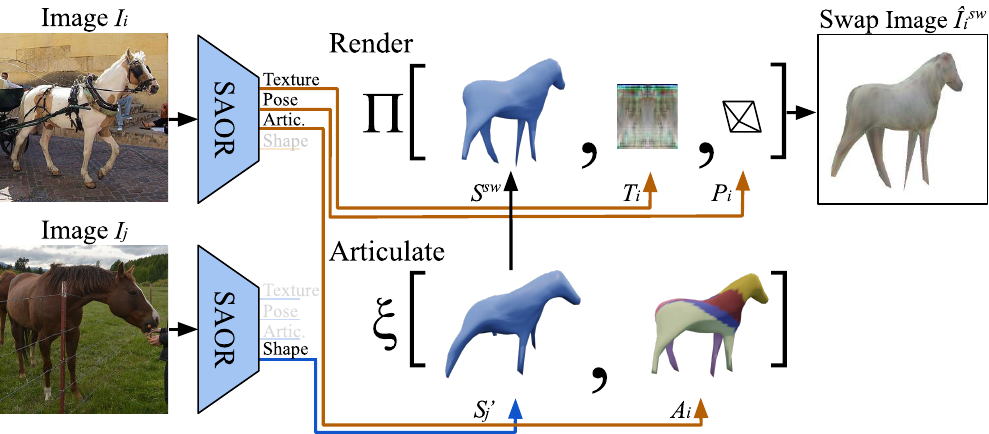}
\vspace{-15pt}
\caption{Illustration of our articulated swap loss. To calculate the loss, a swap image $\hat{I}_{i}^{sw}$ is rendered using a randomly chosen paired image's shape $S_{j}'$, combined with estimated texture, viewpoint, and articulation ($T_{i}, P_{i}, A_{i}$) from the input image $I_{i}$. 
It ensures that 3D predictions are not degenerate and helps disentangle deformation and articulation.}
\label{fig:swap_loss}
\vspace{-10pt}
\end{figure}

One of the hardest challenges in single-view 3D reconstruction is the tendency to predict degenerate solutions as a result of the ill-posed nature of the task (\ie an infinite number of 3D shapes can explain the same 2D input). 
Examples of such failure cases include models predicting flat 2D textured planes which are visually consistent when viewed from the same pose as the input image but lack full 3D shape~\cite{monnier2022share}. 
To mitigate these issues, and to ensure multi-view consistency of our 3D reconstructions, we build on the swap loss idea recently introduced in~\cite{monnier2022share}. 

To estimate their swap loss, \cite{monnier2022share} take a pair of images ($I_{i}, I_{j}$) that depict two different instances of the same object category, and estimate their respective shape, texture, and camera pose, $(\{S_{i}, T_{i}, P_{i}\}$, $\{S_{j}, T_{j}, P_{j}\})$. 
They then generate an image $\hat{I}_i^{sw} = \Pi (S_{j}, T_{i}, P_{i})$ by swapping the shape encodings $S_{i}$ and $S_{j}$, where $\Pi$ is a differentiable renderer. 
Finally, they estimate the appearance loss between $I_{i}$ and $\hat{I}_i^{sw}$ which aims to enforce cross-instance consistency. 
The intuition here is that the shape from $I_{j}$ and texture from $I_{i}$ should be sufficient to describe the appearance of $I_{i}$, even though $I_{j}$ is potentially captured from a different viewpoint. 

In~\cite{monnier2022share}, the shapes $S_{i}$ and $S_{j}$ should be similar, while the predicted viewpoints $P_{i}$ and $P_{j}$ should be different to get a useful `multi-view' training signal. 
To obtain similar shapes, they store latent shape codes in a memory bank which is queried online via a nearest neighbor lookup. 
This memory bank is updated at each iteration for the selected shape codes using the current state of the network. 
Moreover, they limit the search neighborhood based on the predicted viewpoints to ensure that they obtain some viewpoint variation, \ie in~\cite{monnier2022share} the viewpoints $P_{i}$ and $P_{j}$ should not be too similar, or too different. 
While this results in plausible predictions for mostly rigid categories such as birds and cars, for highly articulated animal categories it can led to degenerate solutions due to more variety in terms of shape appearance, as can be seen in Fig.~\ref{fig:qual_unicorn}. 

\noindent\textbf{Swap Loss.} To address this issue, we introduced a straightforward but more effective swap loss that generalizes to articulated object classes. 
Our hypothesis is that given a set of images that contain a variety of viewpoints exhibiting disentangled deformation and articulation, we can use randomly chosen image pairs to calculate the swap loss. 
Since we model the articulation along with the deformation to obtain the final shape, articulation can be used to explain the difference between shapes. 
In our proposed loss, we swap random deformed shapes $S'_{i}$ and $S'_{j}$ from instances of the same object category, but use the original estimated articulation $S^{sw} = \xi(S'_{j}, A_{i})$ and reconstruct the swap image $\hat{I}_i^{sw}=\Pi(S^{sw}, T_{i}, P_{i})$ to calculate the swap loss $\mathcal{L}_{swap} (I_{i}, \hat{I}_i^{sw})$.
Our loss is illustrated in Fig.~\ref{fig:swap_loss}.

\begin{figure}[t]
\centering
  \includegraphics[width=1.0\columnwidth]{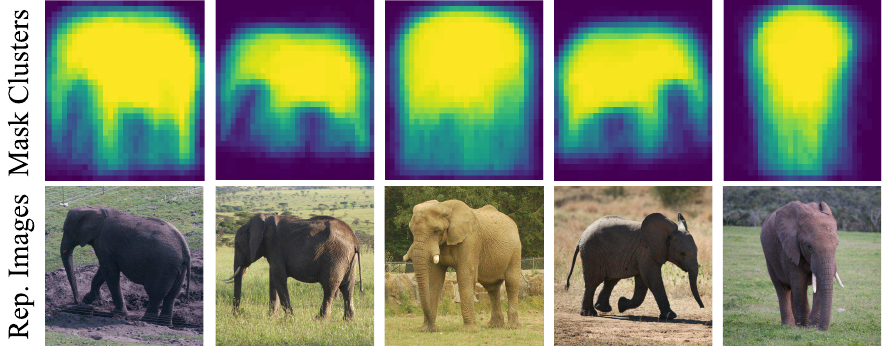}
\vspace{-15pt}
\caption{(Top) Subset of the resulting cluster centers that arise from clustering the object segmentation masks. 
(Bottom) Representative images from each of the clusters above. 
We can see that our simple clustering operation captures the main viewpoint variations present in the data, \eg left facing, frontal, right facing, \etc.}
\label{fig:balanced_sampler}
\vspace{-10pt}
\end{figure}

\noindent\textbf{Balanced Sampling.} For our swap loss to be successful it requires the selected image pairs to ideally be from different viewpoints. 
To obtain informative image pairs, we propose an image sampling mechanism which makes use of the segmentation masks of the input images. 
Before training, we cluster predicted segmentation masks of the training images and then during training we sample images from each cluster uniformly to form batches. 
This ensures that each batch includes the object of interest depicted from different viewpoints. 
In Fig.~\ref{fig:balanced_sampler} we can see that cluster centers mostly capture the rough distribution of viewpoints and thus help stabilize training. 
As our image pairs $(I_i, I_j)$ are sampled from within the same batch during training, this results in varied images from different viewpoints for the swap loss. 
Combined, our swap and balanced sampling steps drastically simplifies the swap loss from~\cite{monnier2022share} and improves reconstruction quality and training stability on articulated classes.

\subsection{Optimization}
Given an input image, $I$, we reconstruct it as $\hat{I}$ using estimated shape, texture, and viewpoint. 
In addition, we use the swapped shape to predict another image $\hat{I}^{sw}$ and calculate the swap loss, as discussed in Sec.~\ref{sec:swap}. 
We also use differentiable rendering to obtain a predicted object segmentation mask and depth derived from the predicted 3D shape, $\hat{M}$ and $\hat{D}$ respectively. 
Our model is trained using a combination of the following losses,
\begin{equation}
 \mathcal{L} = \mathcal{L}_{appr} + \mathcal{L}_{mask} + \mathcal{L}_{depth} +\mathcal{L}_{swap} + \mathcal{L}_{reg}.
\end{equation} 

The appearance loss, $\mathcal{L}_{appr}(I, \hat{I})$, is an RGB and perceptual loss \cite{zhang2018unreasonable}, $\mathcal{L}_{depth}(D, \hat{D})$ is the translation and shift-invariant depth loss introduced in~\cite{Ranftl2021}, and $\mathcal{L}_{mask}(M, \hat{M})$ estimates silhouette discrepancy.
To avoid degenerate solutions, we use $\mathcal{L}_{swap}(I, \hat{I}^{sw})$ and regularize predictions using $\mathcal{L}_{reg}$, which encourages smoothness~\cite{desbrun1999implicit} and normal consistency on the predicted 3D shape along with a uniform distribution on the part assignment. While we use predicted segmentation masks and relative depth during training, at test time, our model only requires a single image.

\subsection{Implementation Details}

We employ a ResNet~\cite{he2016deep} as our global encoder, $f_{enc}$, and perform end-to-end training using Adam~\cite{kingma2014adam}. 
Object masks $M$ and depths $D$ are obtained for training by utilizing off-the-shelf pre-trained networks. 
To implement all 3D operations in our model we use the Pytorch3D framework~\cite{ravi2020accelerating} using their default mesh rasterization~\cite{liu2019soft} which is differentiable and enables end-to-end training. 
Prior to being passed to the model, images are resized to $128\x128$ pixels. 
We disable articulation for the first 100 epochs when training a model from scratch, and continue training models for another 100 epochs by enabling deformation and articulation jointly. 
The lightweight design of our proposed method enables the estimation of the final shape, articulation, texture, and viewpoint in approximately 15 ms per image. 
We provide more details regarding losses, hyperparameters, and optimization in the supplementary material.

\section{Experiments}
\label{sec:exps}

\begin{figure}[t]
    \vspace{-8pt}
    \centering
    \begin{subfigure}{0.15\columnwidth}
    \centering
    \begin{picture}(\columnwidth, 10)
    \put(7,0){Source}
    \end{picture}
    \end{subfigure}
    \centering
    \begin{subfigure}{0.15\columnwidth}
    \centering
    \begin{picture}(\columnwidth, 8)
    \put(5,0){\small{$\lambda=1.0$}}
    \end{picture}
    \end{subfigure}
    \centering
    \begin{subfigure}{0.15\columnwidth}
    \centering
    \begin{picture}(\columnwidth, 8)
    \put(5,0){\small{$\lambda=0.7$}}
    \end{picture}
    \end{subfigure}
    \centering
    \begin{subfigure}{0.15\columnwidth}
    \centering
    \begin{picture}(\columnwidth, 8)
    \put(5,0){\small{$\lambda=0.3$}}
    \end{picture}
    \end{subfigure}
     \centering
    \begin{subfigure}{0.15\columnwidth}
    \centering
    \begin{picture}(\columnwidth, 8)
    \put(5,0){\small{$\lambda=0.0$}}
    \end{picture}
    \end{subfigure}
     \centering
    \begin{subfigure}{0.15\columnwidth}
    \centering
    \begin{picture}(\columnwidth, 8)
    \put(8,0){Target}
    \end{picture}
    \end{subfigure}
    
    \vspace{1.5pt}
    \begin{subfigure}{0.025\columnwidth}
    \centering
    \begin{picture}(\columnwidth, 0)
    \put(0,0){\rotatebox{90}{\scriptsize{Articulation}}}
    \end{picture}
    \end{subfigure}
    \centering
    \begin{subfigure}{0.15\columnwidth}
    \centering
    \includegraphics[width=\columnwidth, trim=10 10 10 10, clip]{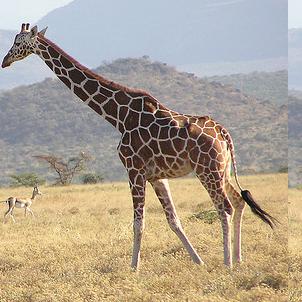}
    \end{subfigure}
    \begin{subfigure}{0.15\columnwidth}
    \centering
    \includegraphics[width=\columnwidth, trim=10 10 10 10, clip]{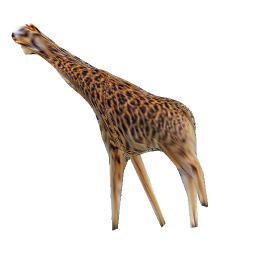}
    \end{subfigure}
    \begin{subfigure}{0.15\columnwidth}
    \centering
    \includegraphics[width=\columnwidth, trim=10 10 10 10, clip]{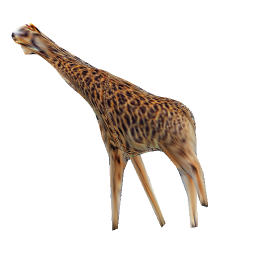}
    \end{subfigure}
    \begin{subfigure}{0.15\columnwidth}
    \centering
    \includegraphics[width=\columnwidth, trim=10 10 10 10, clip]{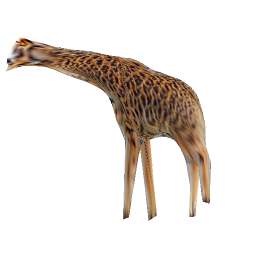}
    \end{subfigure}
    \begin{subfigure}{0.15\columnwidth}
    \centering
    \includegraphics[width=\columnwidth, trim=10 10 10 10, clip]{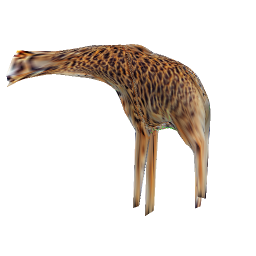}
    \end{subfigure}
    \centering
    \begin{subfigure}{0.15\columnwidth}
    \centering
    \includegraphics[width=\columnwidth, trim=10 10 10 10, clip]{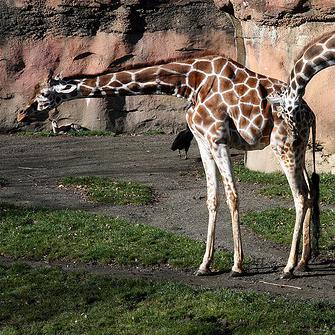}
    \end{subfigure}

   \begin{subfigure}{0.025\columnwidth}
    \centering
    \begin{picture}(\columnwidth, 0)
    \put(0,0){\rotatebox{90}{\scriptsize{Deformation}}}
    \end{picture}
    \end{subfigure}
    \centering
    \begin{subfigure}{0.15\columnwidth}
    \centering
    \includegraphics[width=\columnwidth, trim=10 10 10 10, clip]{figs/art_vs_def_new/source.jpg}
    \end{subfigure}
    \begin{subfigure}{0.15\columnwidth}
    \centering
    \includegraphics[width=\columnwidth, trim=10 10 10 10, clip]{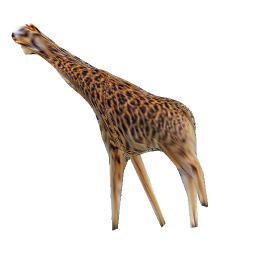}
    \end{subfigure}
    \begin{subfigure}{0.15\columnwidth}
    \centering
    \includegraphics[width=\columnwidth, trim=10 10 10 10, clip]{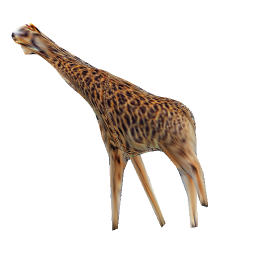}
    \end{subfigure}
    \begin{subfigure}{0.15\columnwidth}
    \centering
    \includegraphics[width=\columnwidth, trim=10 10 10 10, clip]{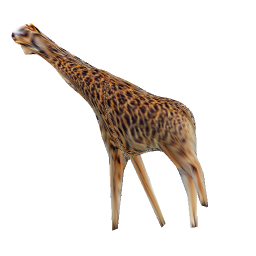}
    \end{subfigure}
    \begin{subfigure}{0.15\columnwidth}
    \centering
    \includegraphics[width=\columnwidth, trim=10 10 10 10, clip]{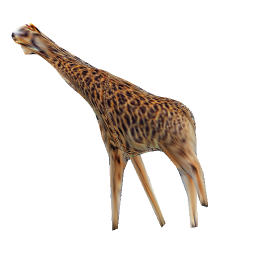}
    \end{subfigure}
    \centering
    \begin{subfigure}{0.15\columnwidth}
    \centering
    \includegraphics[width=\columnwidth, trim=10 10 10 10, clip]{figs/art_vs_def_new/target.jpg}
    \end{subfigure}

\vspace{-5pt}    
\caption{Disentanglement of articulation and deformation. On top, we interpolate articulation latent features between a source and target image, and on the bottom do the same for shape deformation features. $\lambda=1$ indicates that original features are used for reconstruction, while $\lambda=0$ indicates the target ones. 
We can see that the difference between the reconstructions is explained by articulation changes between the source and target image pairs.
} 
\label{fig:art_vs_def}
\vspace{-10pt}
\end{figure}

Here we present results on multiple quadruped and biped animal categories, providing both quantitative and qualitative comparisons to previous work. 

\subsection{Data and Pre-Processing}

For our experiments, we trained two models: \textbf{SAOR-Bird} and \textbf{SAOR-101}. The bird model is trained from scratch using the CUB~\cite{CUB} dataset following the original train/test split. SAOR-101, the general animal model, is trained on 101 animal categories that contain birds, quadrupeds, and bipeds. This model is first trained using only horse images from the LSUN~\cite{yu2015lsun} dataset with an additional 500 front-facing horse images from iNaturalist \cite{iNatWeb}, as LSUN mostly contains side-view images of horses. 
Then, as in~\cite{wu2022magicpony}, we finetune the horse model on a new dataset that we collected from iNaturalist~\cite{iNatWeb} which contains 90k images from 101 different animal classes. In the supplementary material, we provide more details about the dataset.

For pre-processing, we run a general-purpose animal object detector~\cite{beery2019efficient} to detect all the animals present in the input images and then filter the detections based on the confidence, size, and location of the bounding box. 
We then extract segmentation masks using SAM~\cite{kirillov2023segany} and estimate the relative monocular depth using MiDaS~\cite{Ranftl2021,Ranftl2022}. 

\subsection{Quantitative Results}
To compare to existing work, we quantitatively evaluate using the 2D keypoint transfer task, which reflects the quality of the estimated shape and viewpoint and 3D evaluation which reflects how predicted and ground truth depth is aligned 
We report results using the PCK metric with a 0.1 threshold for the keypoint transfer task and normalized L1 Chamfer distance for 3D evaluation.

\noindent\textbf{Birds.} Keypoint transfer results on CUB~\cite{CUB} are presented in Table~\ref{tab:pck_cub}, both for all bird classes and the non-aquatic subset as in~\cite{wu2022magicpony}. 
Our method obtains the best results out of methods that do not use keypoint supervision, 3D object priors (\eg 3D templates or skeletons~\cite{wu2021dove,wu2022magicpony}), or additional data (\eg \cite{wu2021dove,wu2022magicpony}). 

\begin{table}
  \centering
  \resizebox{0.9\columnwidth}{!}{
  \begin{tabular}{l l c c }
  \toprule
     Supervision & Method & all & w/o aqua \\ \midrule
     \stemp\newstar, \smask, \skey, \sview & CMR~\cite{kanazawa2018learning} & 54.6 & 59.1 \\
     \stemp\newstar, \smask& U-CMR~\cite{goel2020shape} & 35.9 & 41.2 \\ 
     \skel, \smask\newstar,\svid, \sflow \newstar & DOVE \cite{wu2021dove} & 44.7 & 51.0 \\ 
     \skel, \smask\newstar,\sssl, $\dagger$ & MagicPony~\cite{wu2022magicpony} & 55.5 & 63.5 \\ \midrule
     \smask & CMR~\cite{kanazawa2018learning} & 25.5 & 27.7 \\
     \smask, SCOPS\newstar & UMR~\cite{li2020self} & 51.2 & 55.5 \\
    None & Unicorn \cite{monnier2022share}& 49.0 & 53.5 \\ \midrule
    \smask \newstar, \sdepth \newstar & SAOR-Bird & 51.9 & 57.8 \\ \bottomrule
  \end{tabular}
  }
  \vspace{-5pt}
  \caption{Keypoint transfer results on CUB~\cite{CUB} using the PCK metric with 0.1 threshold (higher is better).
  \stemp~3D template shape, \skel~3D skeleton, \sview~camera viewpoint, \skey~2D keypoints, \smask~segmentation mask, \sflow~optical flow, \svid~video, \sssl~DINO features, SCOPS part segmentation, and \sdepth~monocular depth.
  $\dagger$~also uses additional video frames from \cite{wu2021dove}. 
  The initial 3D template in \cite{kanazawa2018learning,goel2020shape} is derived from 2D keypoints. 
  \newstar~indicates that the supervision is predicted, hence it is weak supervision. We obtain the best results for methods that do not use 3D templates (\stemp), skeletons (\skel), or extra data during training in addition to CUB (\eg \cite{wu2021dove,wu2022magicpony}).
  }
  \label{tab:pck_cub}
  \vspace{-5pt}
\end{table}

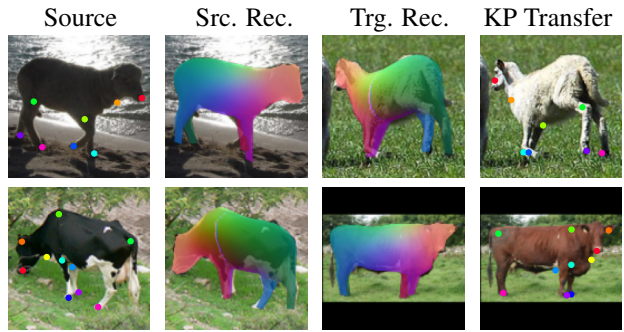
\begin{figure}[t]
    \vspace{-6pt}
    \centering
    \begin{subfigure}{0.24\columnwidth}
    \centering
    \begin{picture}(\columnwidth, 10)
    \put(15,2){Source}
    \end{picture}
    \end{subfigure}
    \centering
    \begin{subfigure}{0.24\columnwidth}
    \centering
    \begin{picture}(\columnwidth, 8)
    \put(13,2){Src. Rec.}
    \end{picture}
    \end{subfigure}
    \centering
    \begin{subfigure}{0.24\columnwidth}
    \centering
    \begin{picture}(\columnwidth, 8)
    \put(12,2){Trg. Rec.}
    \end{picture}
    \end{subfigure}
    \centering
    \begin{subfigure}{0.24\columnwidth}
    \centering
    \begin{picture}(\columnwidth, 8)
    \put(3,2){KP Transfer}
    \end{picture}
    \end{subfigure}
    \vspace{-4pt}
    \centering
    \begin{subfigure}{\columnwidth}
    \centering
    \includegraphics[width=\columnwidth, trim=10 10 10 27, clip]{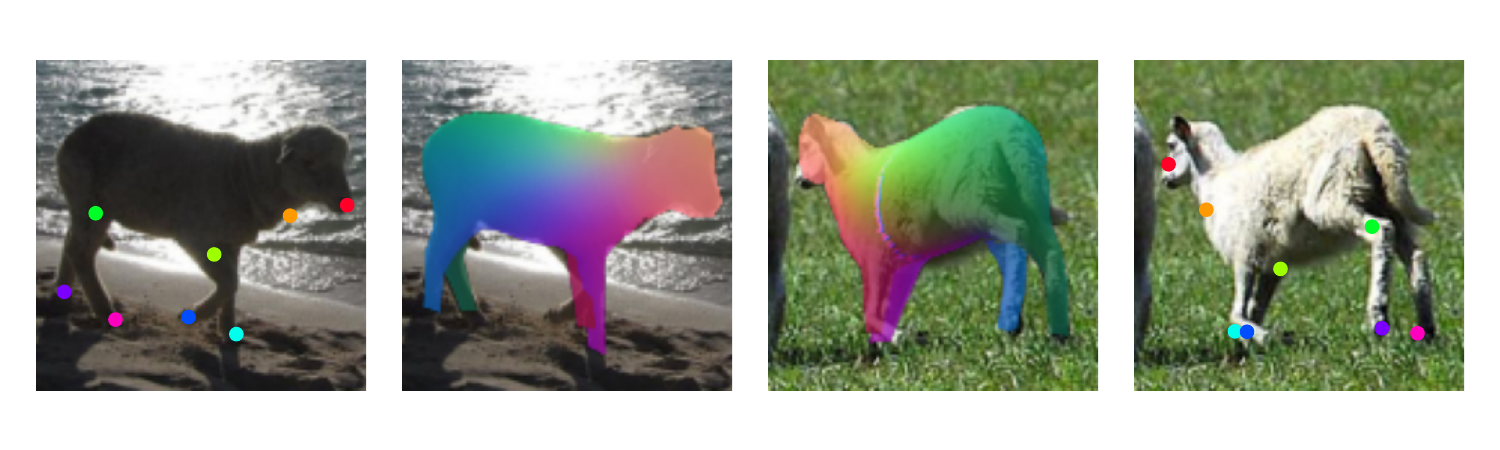}
    \end{subfigure}
    \centering
    \vspace{-8pt}
    \centering
    \begin{subfigure}{\columnwidth}
    \centering
    \includegraphics[width=\columnwidth, trim=10 10 10 27, clip]{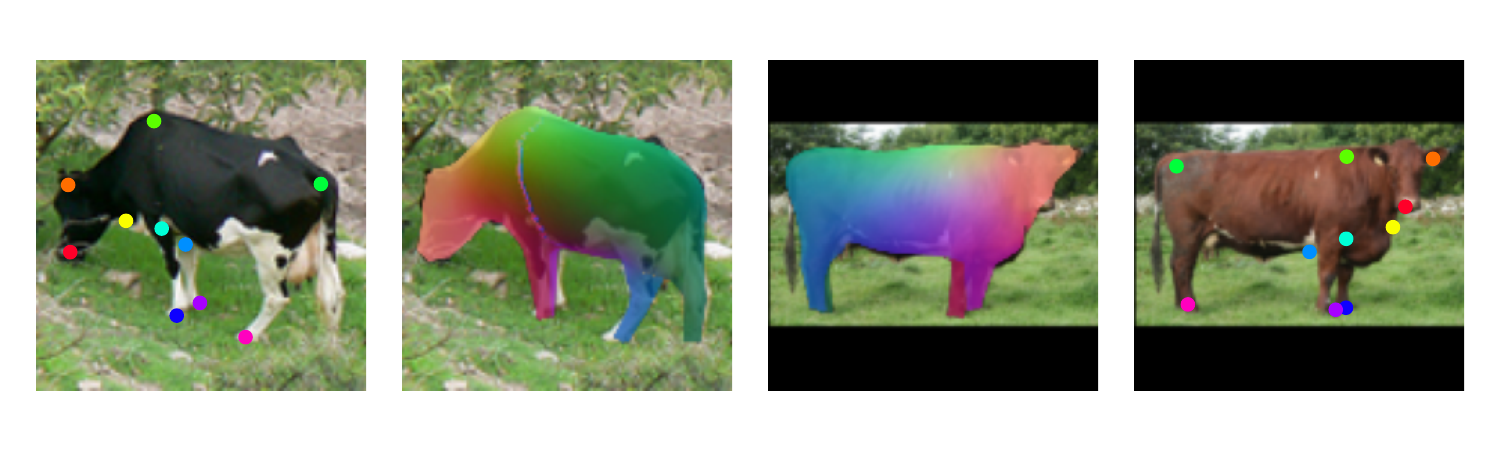}
    \end{subfigure}
\vspace{-16pt}    
\caption{Keypoint transfer results. Our model captures articulation and viewpoint differences between images.} 
\label{fig:kp}
\vspace{-6pt}
\end{figure}

\noindent\textbf{Quadrupeds.} Keypoint transfer results for quadruped animals from the Pascal dataset~\cite{everingham2015pascal} are presented in Table~\ref{tab:pck_animals}. 
As noted earlier, we trained the horse model from scratch, while the other models were finetuned using data from iNaturalist~\cite{iNatWeb}. 
For the Unicorn~\cite{monnier2022share} baseline, we used their pre-trained model which was also trained on LSUN horses. 
For the remaining categories, we also finetuned their model in a similar fashion to ours.
Our method outperforms CSM~\cite{kulkarni2019csm} and its articulated version A-CSM~\cite{kulkarni2020acsm}, which use a 3D template of the object category and 3D part segmentation for the horse and cow category. 
Moreover, our method achieved significantly better scores than Unicorn~\cite{monnier2022share}, which produces degenerate (\ie flat) shape predictions for these classes (see Fig.~\ref{fig:qual_unicorn}).
We visualize some keypoint transfer results in Fig~\ref{fig:kp}.

\begin{table}
  \centering
  \resizebox{0.9\columnwidth}{!}{
  \begin{tabular}{l l c r r}
  \toprule
     Supervision & Method & Horse & Cow & Sheep \\ \midrule
     \smask & Dense-Equi~\cite{thewlis2017unsupervisedneurips} & 23.3 & 20.9 & 19.6 \\
     \smask, \stemp & CSM~\cite{kulkarni2019csm} & 31.2 & 26.3 & 24.7\\
     \smask, \stemp & A-CSM~\cite{kulkarni2020acsm} & 32.9 & 26.3 & 28.6\\
      \midrule
      None & Unicorn~\cite{monnier2022share} & 14.9 & 12.1 & 11.0 \\
      \skel, \smask\newstar,\sssl, $\dagger$ & MagicPony~\cite{wu2022magicpony} & 42.9 & 42.5 & 26.2 \\
     \midrule
     \smask \newstar , \sdepth \newstar & SAOR-101 & 44.9 & 33.6 & 29.1 \\

     \bottomrule
  \end{tabular}
  }
  \vspace{-5pt}
  \caption{Keypoint transfer results for quadruped animals.}
  \label{tab:pck_animals}
  \vspace{-5pt}
\end{table}

We also present 3D evaluation using results using Animal3D dataset~\cite{xu2023animal3d} on a few quadruped categories in Table~\ref{tab:animal_3d_chamfer}. We calculate the normalized L1 Chamfer distance between ground truth and predictions after running ICP~\cite{besl1992method} between the ground truth and predictions as there is no canonical alignment between methods and ground truth data. SAOR obtains better results than Unicorn~\cite{monnier2022share} and similar results to MagicPony~\cite{wu2022magicpony}, while being category agnostic. 

\begin{table}
  \centering
  \resizebox{0.9\columnwidth}{!}{
  \begin{tabular}{c l c r r}
  \toprule
     Supervision & Method & Horse & Cow & Sheep \\ \midrule
     None & Unicorn~\cite{monnier2022share} & 0.091 & 0.118 & 0.134 \\
    \skel, \smask\newstar,\sssl, $\dagger$ & MagicPony~\cite{wu2022magicpony} & 0.046 & 0.040 & - \\ 
     \smask \newstar , \sdepth \newstar & SAOR-101 & 0.046 & 0.043 & 0.045 \\
     \bottomrule
  \end{tabular}
  }
  \vspace{-5pt}
  \caption{3D evaluation on the Animal3D dataset~\cite{xu2023animal3d} using normalized L1 Chamfer error, where lower is better. }
  \vspace{-8pt}
  \label{tab:animal_3d_chamfer}
\end{table}

\subsection{Ablation Experiments}
To provide insight into the impact of our proposed model components, we provide ablation experiments on Pascal for quadrupeds and on CUB for birds in Table~\ref{tab:ablation}. 
While depth information helps to improve results, we can see that our articulation and swap modules are significantly more important. 
Our model trained without the swap loss obtains reasonable keypoint matching performance for birds but produces degenerate flat plane-like solutions and fails miserably for quadrupeds. 
The performance also drops if articulation is not utilized. 
This is because we choose random pairs for the swap loss (unlike \cite{monnier2022share}'s more expensive pair selection), and thus only viewpoint changes can be used to explain the difference between images.

\begin{table}
  \centering
  \resizebox{0.9\columnwidth}{!}{
  \begin{tabular}{l c c c c}
  \toprule
     Method & Horse & Cow & Sheep & Bird\\ \midrule
     Ours & 44.9 & 33.6 & 29.1 & 51.9 \\
     Ours w/o depth & 42.4 & 30.1 & 26.8 & 49.9 \\
     Ours w/o swap & 30.8 & 17.7 & 18.4 & 44.5 \\
     Ours w/o sampling & 27.5 & 20.1 & 18.3 & 38.8 \\
     Ours w/o articulation & 26.3 & 19.4 & 17.9 & 41.7 \\
     \bottomrule
  \end{tabular}
  }
  \vspace{-5pt}
  \caption{Keypoint transfer ablation results for SAOR where we
  disable individual components to measure their impact. 
  }
  \vspace{-5pt}
  \label{tab:ablation}
\end{table}

\subsection{Qualitative Results}
\begin{figure}[t]

\centering
    \centering
    
    \begin{subfigure}{\columnwidth}
    \begin{picture}(\columnwidth, 5)
    \put(0.05\columnwidth,0){Input}
    \put(0.35\columnwidth,0){SAOR-101}
    \put(0.7\columnwidth,0){Unicorn~\cite{monnier2022share}}
    \end{picture}
    \end{subfigure}
    
   \centering
    \begin{subfigure}{0.20\columnwidth}
    \centering
    \includegraphics[width=\columnwidth, trim=0 0 0 0, clip]{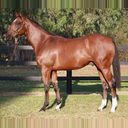}
    \end{subfigure}
    \centering
    \begin{subfigure}{0.19\columnwidth}
    \centering
    \includegraphics[width=\columnwidth, trim=0 0 0 0, clip]{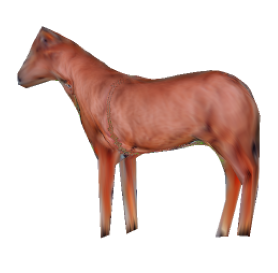}
    \end{subfigure}
    \begin{subfigure}{0.19\columnwidth}
    \centering
    \includegraphics[width=\columnwidth, trim=0 0 0 0, clip]{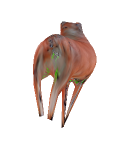}
    \end{subfigure}
    \centering
    \begin{subfigure}{0.19\columnwidth}
    \centering
    \includegraphics[width=\columnwidth, trim=0 0 0 0, clip]{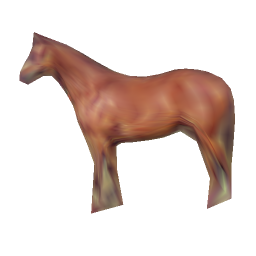}
    \end{subfigure}
    \centering
    \begin{subfigure}{0.19\columnwidth}
    \centering
    \includegraphics[width=\columnwidth, trim=0 0 0 0, clip]{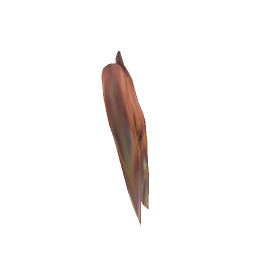}
    \end{subfigure}
    \begin{subfigure}{\columnwidth}
    \begin{picture}(\columnwidth, 3)
    \put(0.05\columnwidth,0){}
    \put(0.35\columnwidth,0){SAOR-101}
    \put(0.70\columnwidth,0){UMR~\cite{li2020self}}
    \end{picture}
    \end{subfigure}
    \centering
    \begin{subfigure}{0.21\columnwidth}
    \centering
    \includegraphics[width=\columnwidth, trim=0 0 0 0, clip]{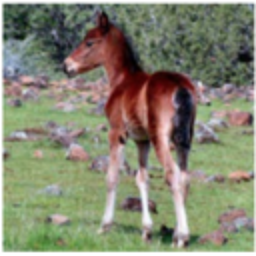}
    \end{subfigure}
    \centering
    \begin{subfigure}{0.21\columnwidth}
    \centering
    \includegraphics[width=\columnwidth, trim=0 0 0 0, clip]{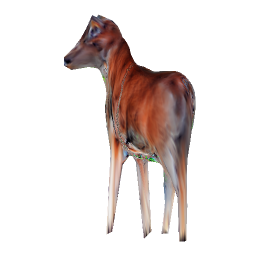}
    \end{subfigure}
    \begin{subfigure}{0.21\columnwidth}
    \centering
    \includegraphics[width=\columnwidth, trim=0 0 0 0 clip]{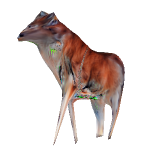}
    \end{subfigure}
    \centering
    \begin{subfigure}{0.33\columnwidth}
    \centering
    \includegraphics[height=1.7cm, trim=0 0 0 0, clip]{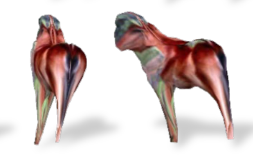}
    \end{subfigure}
\vspace{-8pt}
\caption{Comparison of our model to Unicorn~\cite{monnier2022share} and UMR~\cite{li2020self} on horses. 
Compared to UMR which predicts thin shapes with two legs, we can reconstruct multi-view consistent results with four legs. 
Unicorn fails to produce 3D consistent shapes.}
\label{fig:qual_unicorn}
\vspace{-5pt}
\end{figure}
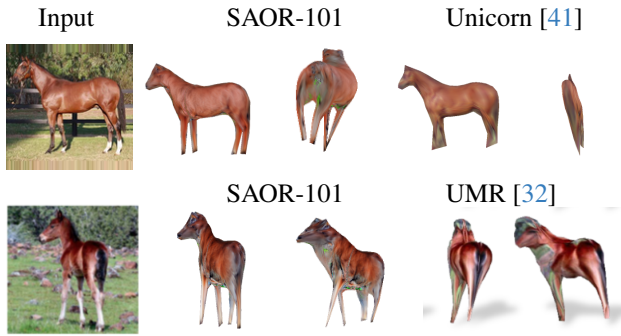

\begin{figure}[t]
    \begin{subfigure}{\columnwidth}
    \begin{picture}(\columnwidth, 5)
    \put(0.05\columnwidth,0){Input}
    \put(0.34\columnwidth,0){SAOR-101}
    \put(0.67\columnwidth,0){MagicPony~\cite{wu2022magicpony}}
    \end{picture}
    \end{subfigure}
    
    \begin{subfigure}{0.19\columnwidth}
    \includegraphics[width=\columnwidth, trim=0 0 0 0, clip]{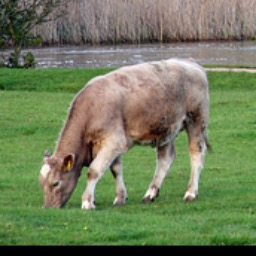}
    \end{subfigure}
    \begin{subfigure}{0.19\columnwidth}
    \includegraphics[width=\columnwidth, trim=0 0 0 0, clip]{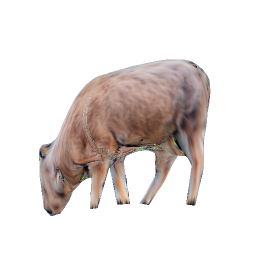}
    \end{subfigure}
    \begin{subfigure}{0.19\columnwidth}
    \includegraphics[width=\columnwidth, trim=50 50 50 50, clip]{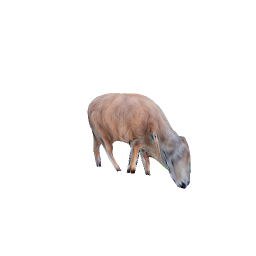}
    \end{subfigure}
    \begin{subfigure}{0.19\columnwidth}
    \includegraphics[width=\columnwidth, trim=0 0 0 0, clip]{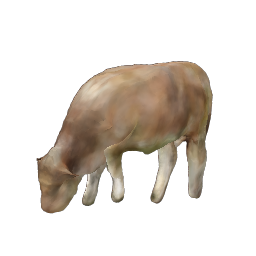}
    \end{subfigure}
    \begin{subfigure}{0.19\columnwidth}
    \includegraphics[width=\columnwidth, trim=0 0 30 0, clip]{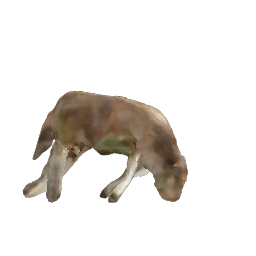}
    \end{subfigure}
    
    \begin{subfigure}{0.19\columnwidth}
    \includegraphics[width=\columnwidth, trim=0 0 0 0, clip]{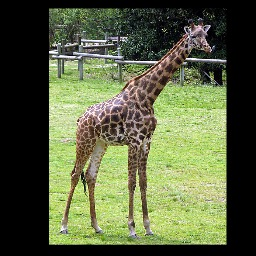}
    \end{subfigure}
    \begin{subfigure}{0.19\columnwidth}
    \includegraphics[width=\columnwidth, trim=0 0 0 0 , clip]{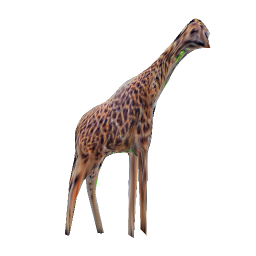}
    \end{subfigure}
     \begin{subfigure}{0.19\columnwidth}
    \includegraphics[width=\columnwidth, trim=50 50 50 50 , clip]{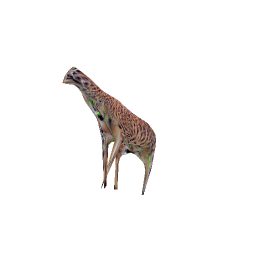}
    \end{subfigure}
    \begin{subfigure}{0.19\columnwidth}
    \includegraphics[width=\columnwidth, trim=0 0 0 0, clip]{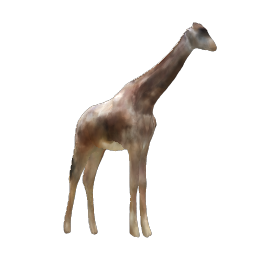}
    \end{subfigure}
    \begin{subfigure}{0.19\columnwidth}
    \includegraphics[width=\columnwidth, trim=40 40 40 40, clip]{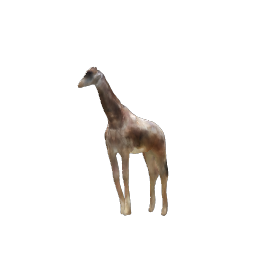}
    \end{subfigure}

    \vspace{-8pt}
    \begin{subfigure}{0.19\columnwidth}
    \includegraphics[width=\columnwidth, trim=0 0 0 0, clip]{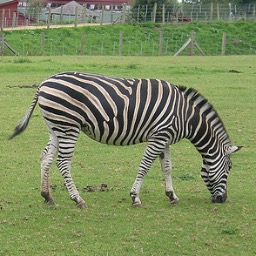}
    \end{subfigure}
    \begin{subfigure}{0.19\columnwidth}
    \includegraphics[width=\columnwidth, trim=0 0 0 0, clip]{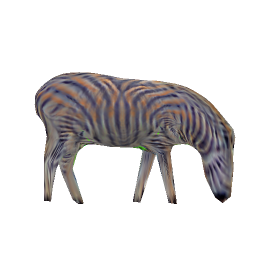}
    \end{subfigure}
     \begin{subfigure}{0.19\columnwidth}
    \includegraphics[width=\columnwidth, trim=50 50 50 50, clip]{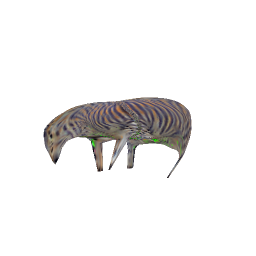}
    \end{subfigure}
     \begin{subfigure}{0.19\columnwidth}
    \includegraphics[width=\columnwidth, trim=0 0 0 0, clip]{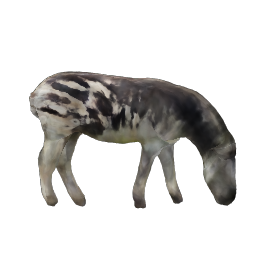}
    \end{subfigure}
     \begin{subfigure}{0.19\columnwidth}
    \includegraphics[width=\columnwidth, trim=20 0 20 0, clip]{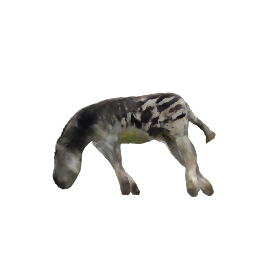}
    \end{subfigure}
    
\vspace{-8pt}
\caption{Comparison of our model to MagicPony~\cite{wu2022magicpony} (without texture refinement) which uses a  category specific skeleton prior during training. 
We obtain on-par reconstructions compared to MagicPony without using any 3D prior on the articulation of the object class and with a simpler and more efficient architecture. }
\label{fig:qual_saor_vs_mp}
\vspace{-14pt}
\end{figure}
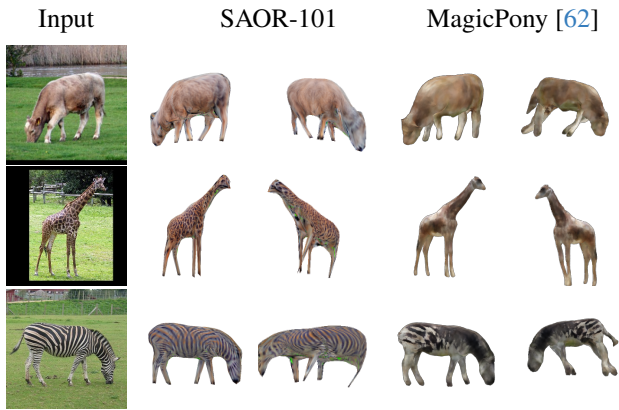

\vspace{-5pt}
\noindent\textbf{Comparison with Previous Work.} We compare SAOR with methods that do not use any 3D shape priors (\ie Unicorn~\cite{monnier2022share} and UMR~\cite{li2020self}) and methods that use a 3D skeleton prior (\ie MagicPony~\cite{wu2022magicpony}). 
A comparison of shape predictions for horses can be seen in Figs.~\ref{fig:qual_unicorn} and~\ref{fig:qual_saor_vs_mp}. 
While Unicorn produces reasonable reconstructions from the input viewpoint, their predictions are flat from the side. 
UMR also predicts thin 3D shapes and does not generate four legs. 
Our method reconstructs multi-view consistent 3D shapes, with prominent four legs. 
In general, our method produces similar results to MagicPony. 
However, MagicPony's hybrid volumetric-mesh representation requires an extra transformation from implicit to explicit representation using~\cite{shen2021deep} and requires multiple rendering operations to estimate the final shape. Moreover, the texture predictions of our methods do not require test-time optimization.

\noindent\textbf{Deformation and Articulation Disentanglement.} In Fig.~\ref{fig:art_vs_def} we illustrate the disentanglement of articulation and deformation learned by our model. 
Given two images depicting differently articulating instances, we interpolate the deformation and articulation features between them to visualize reconstructions. 
While interpolating the articulation feature changes the result, changing the deformation feature does not as the shape difference between both images can be explained via articulation changes.

\noindent\textbf{Part Consistency.} After finetuning the pre-trained horse model on different quadruped categories, we observe that the predicted part assignments stay consistent across categories, as can be seen in Fig.~\ref{fig:teaser}. 
For instance, although the shapes of giraffes and elephants are significantly different, our method is able to assign similar parts to similarly articulated areas. 
Here, each color represents the part that is predicted with the highest probability from the part assignment matrix $W$ by the articulation network $f_{a}$.

\begin{figure}[t]
    \vspace{-4pt}
    \centering
        \begin{subfigure}{0.48\columnwidth}
        \centering
        \includegraphics[width=\columnwidth, trim=10 10 10 10, clip]{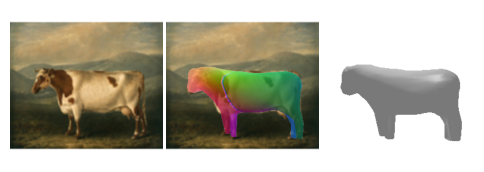}
        \end{subfigure}
        \begin{subfigure}{0.48\columnwidth}
        \centering
        \includegraphics[width=\columnwidth, trim=10 10 10 10, clip]{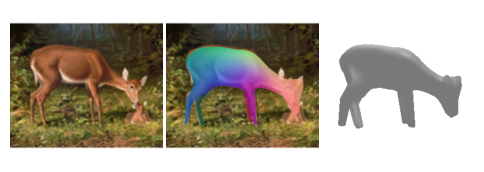}
        \end{subfigure}

        \begin{subfigure}{0.48\columnwidth}
        \centering
        \includegraphics[width=\columnwidth, trim=10 10 10 10, clip]{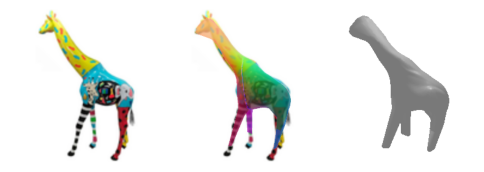}
        \end{subfigure}
        \begin{subfigure}{0.48\columnwidth}
        \centering
        \includegraphics[width=\columnwidth, trim=10 10 10 10, clip]{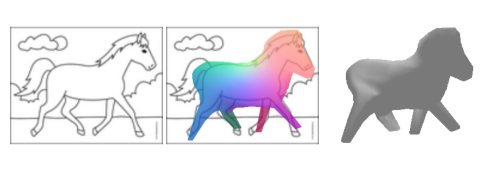}
        \end{subfigure}
        
\vspace{-8pt}    
\caption{Our model, trained on real-world images,  plausibly estimates 3D shape and viewpoint from different domains, \eg cartoons, line drawings, and paintings.} 
\label{fig:gen}
\vspace{-16pt}
\end{figure}

\noindent\textbf{Out-of-Distribution Images.} We illustrate the generalization capabilities of our model by predicting 3D shapes from non-photoreal images, \eg drawings. 
Fig.~\ref{fig:gen} shows that we can reconstruct plausible shapes and poses from input images that are very different from the training domain. 

\vspace{-5pt}
\subsection{Discussion and Limitations}
\vspace{-5pt}
Although our proposed approach is able to estimate plausible 3D shapes, the texture predictions are still not fully realistic. 
This could be improved using test-time refinement similar to~\cite{wu2022magicpony} or alternative texture representations.
During training, our method uses estimated silhouettes and relative depth maps as supervision. 
Both depth maps and silhouettes come from a generic pre-trained models~\cite{Ranftl2022,kirillov2023segany}, hence are free to acquire.
Finally, our method fails to predict accurate shape if the input images contains unusual viewpoints that differ significantly from the training images or the object is not full visible. We present some examples of these failure cases in the supplementary material.

\vspace{-7pt}
\section{Conclusion}
\label{sec:conclusion}
\vspace{-5pt}
We presented SAOR, a new approach for single-view articulated object reconstruction. 
SAOR is capable of predicting the 3D shape of articulated object categories without requiring any explicit object-specific 3D information, \eg 3D templates or skeletons, at training time. 
To achieve this, we learn to segment objects into parts which move together and propose a new swap-based regularization loss that improves 3D shape consistency in addition to simplifying training compared to competing methods. 
These contributions enable us to simultaneously represent over 100 different categories, with diverse shapes, in one model.

\vspace{2pt}
{
\noindent{\bf Acknowledgments:} OMA was in part supported by university partner contributions to the Alan Turing Institute. 
}

{\small
\bibliographystyle{ieee_fullname}
\bibliography{main}
}

\clearpage
\appendix
\setcounter{table}{0}
\renewcommand{\thetable}{A\arabic{table}}
\setcounter{figure}{0}
\renewcommand{\thefigure}{A\arabic{figure}}

\section{Additional Results}

\begin{figure*}
\begin{center}    

    \begin{picture}(0.49\textwidth, 5)
    \put(0.03\textwidth,0){Input}
    \put(0.12\textwidth,0){Pose}
    \put(0.24\textwidth,0){Reconstruction}
    \put(0.41\textwidth,0){Parts}
    \end{picture}
    \begin{picture}(0.49\textwidth, 5)
    \put(0.035\textwidth,0){Input}
    \put(0.13\textwidth,0){Pose}
    \put(0.24\textwidth,0){Reconstruction}
    \put(0.41\textwidth,0){Parts}
    \end{picture}
    \vspace{-5pt}
    \includegraphics[width=0.4975\textwidth, trim=5 0 5 0, clip]{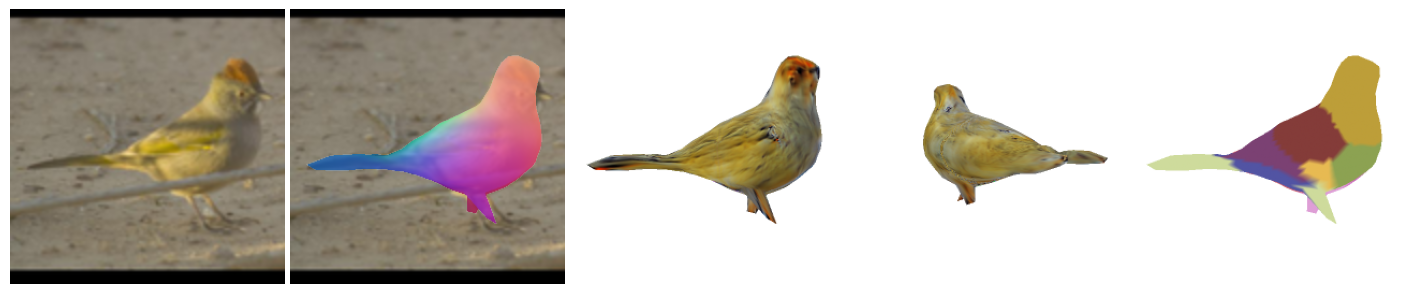}
    \includegraphics[width=0.4975\textwidth, trim=5 0 5 0, clip]{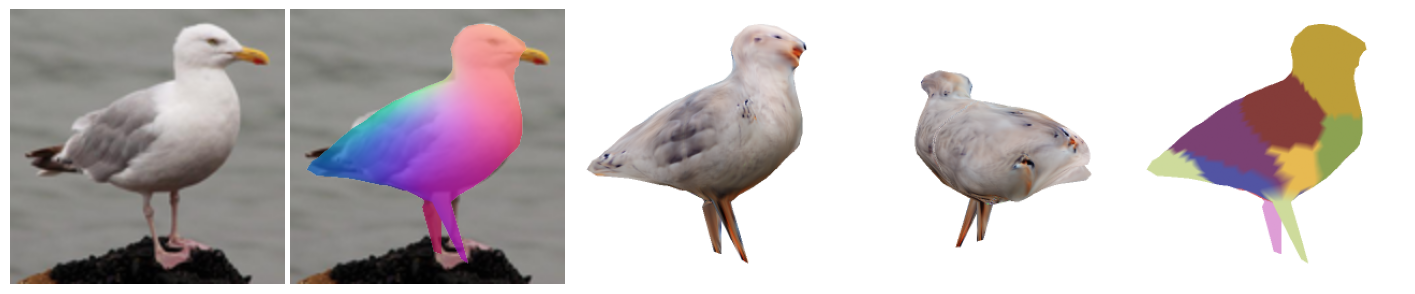}
    \vspace{-5pt}
    \includegraphics[width=0.4975\textwidth, trim=5 0 5 0, clip]{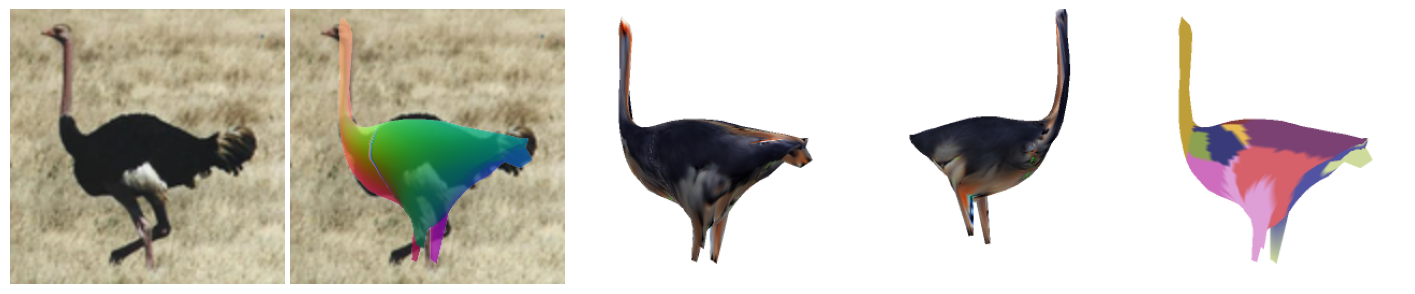}
    \includegraphics[width=0.4975\textwidth, trim=5 0 5 0, clip]{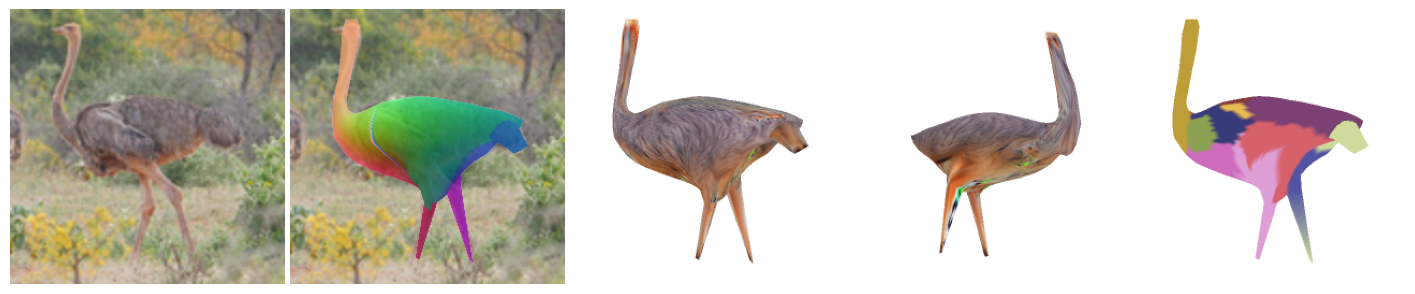}
    \vspace{-5pt}
    \includegraphics[width=0.4975\textwidth, trim=5 0 5 0, clip]{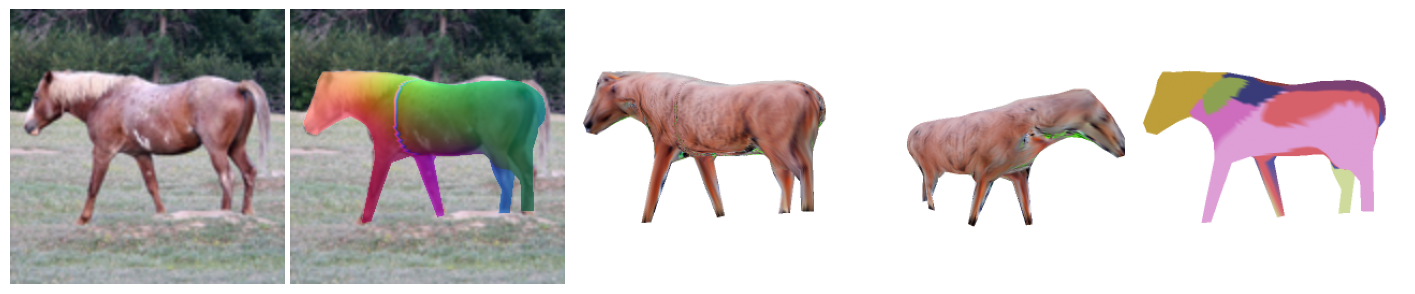}
    \includegraphics[width=0.4975\textwidth, trim=5 0 5 0, clip]{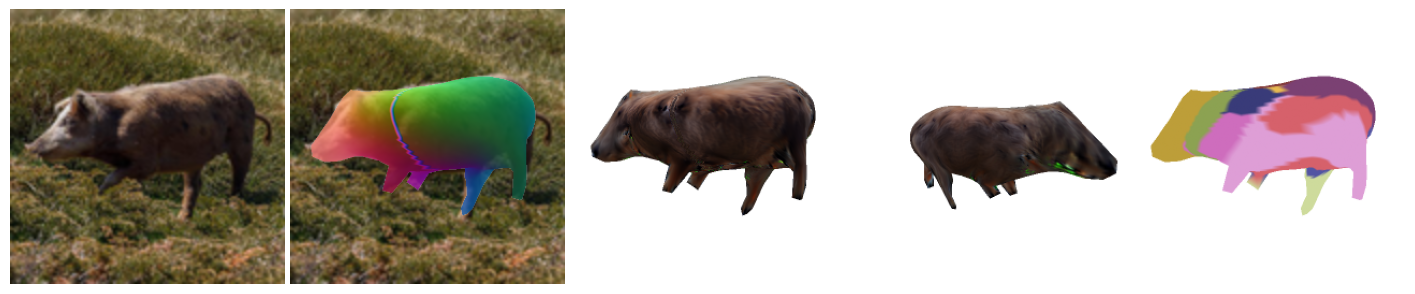}
    \vspace{-5pt}
    \includegraphics[width=0.4975\textwidth, trim=5 0 5 0, clip]{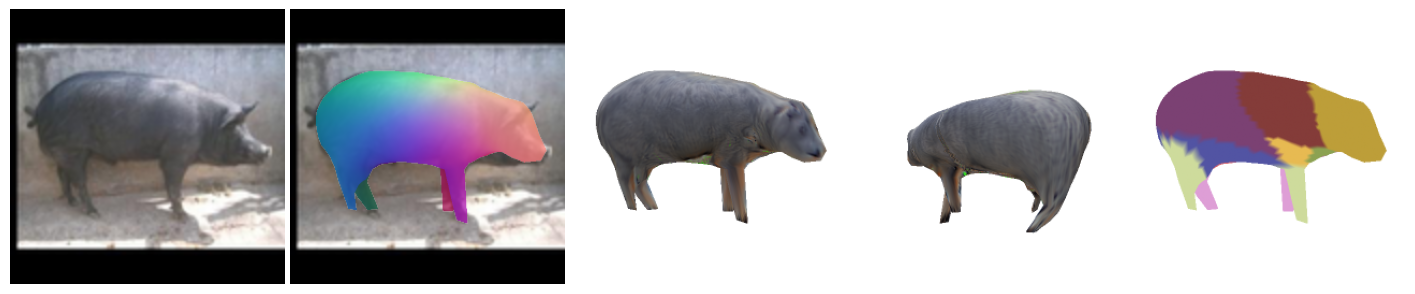}
    \includegraphics[width=0.4975\textwidth, trim=5 0 5 0, clip]{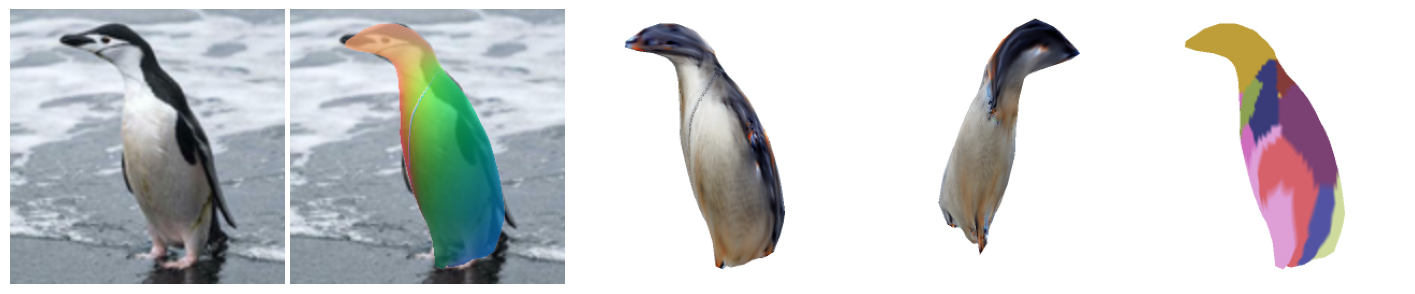}
    \vspace{-5pt}
    \includegraphics[width=0.4975\textwidth, trim=5 0 5 0, clip]{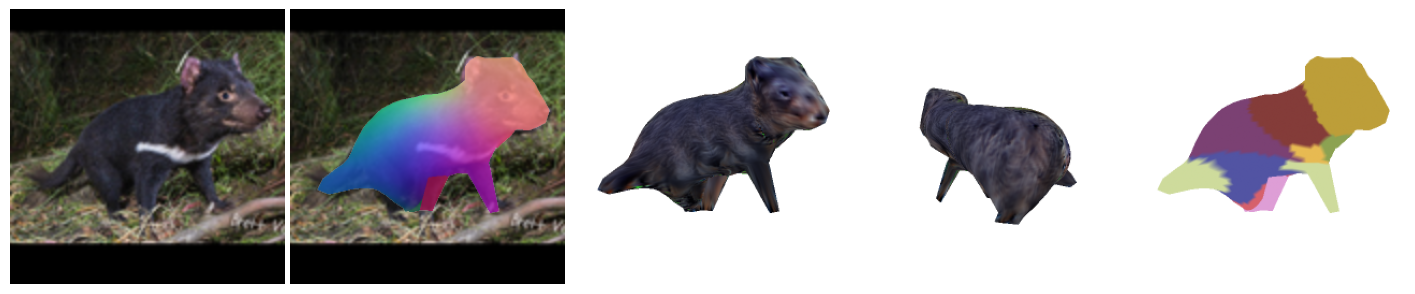}
    \includegraphics[width=0.4975\textwidth, trim=5 0 5 0, clip]{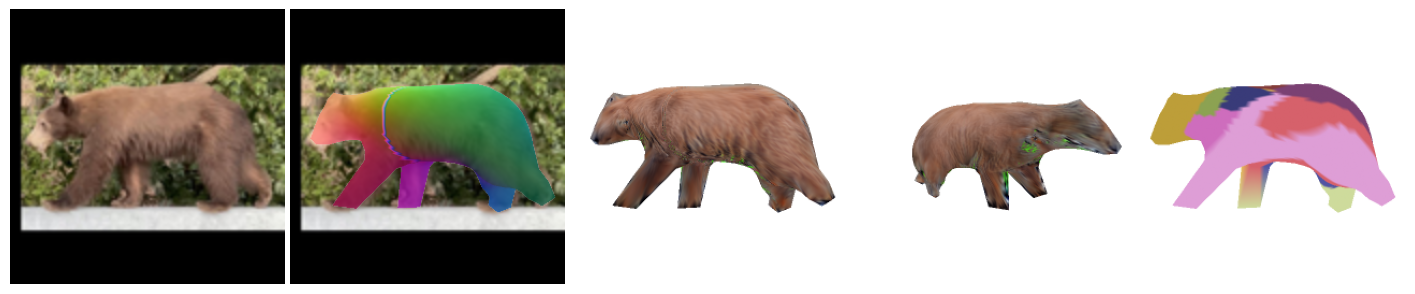}
    \vspace{-5pt}
    \includegraphics[width=0.4975\textwidth, trim=5 0 5 0, clip]{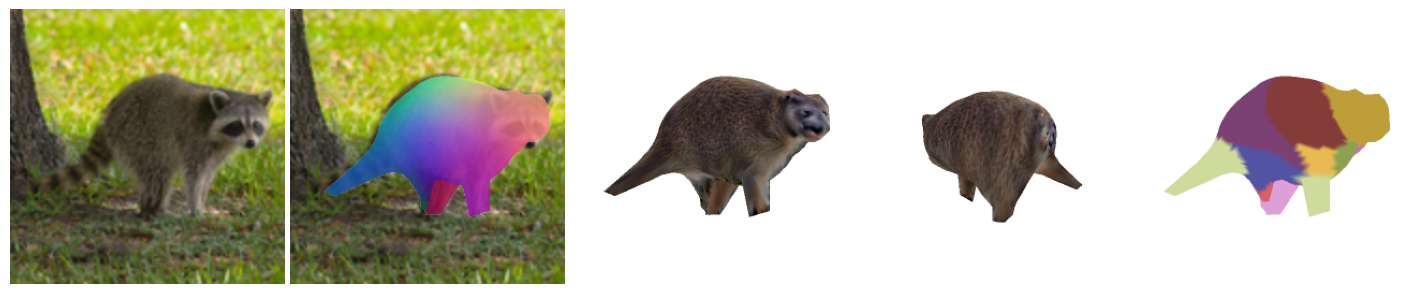}
    \includegraphics[width=0.4975\textwidth, trim=5 0 5 0, clip]{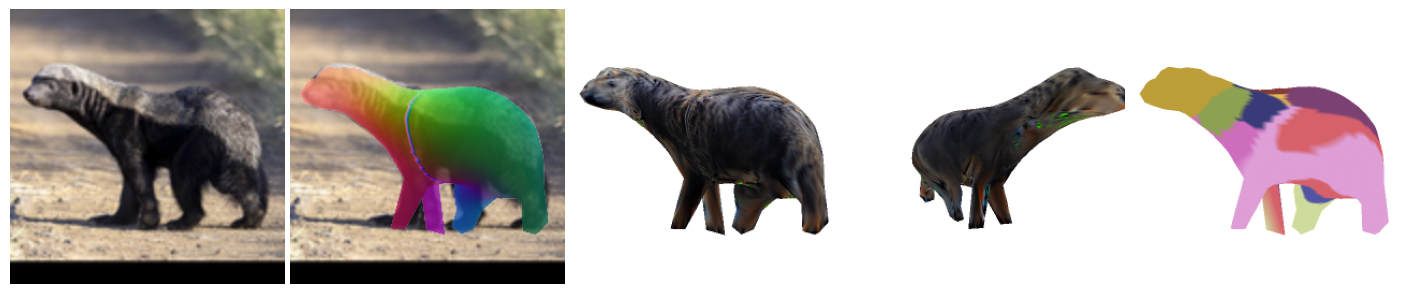}
    \vspace{-5pt}
    \includegraphics[width=0.4975\textwidth, trim=5 0 5 0, clip]{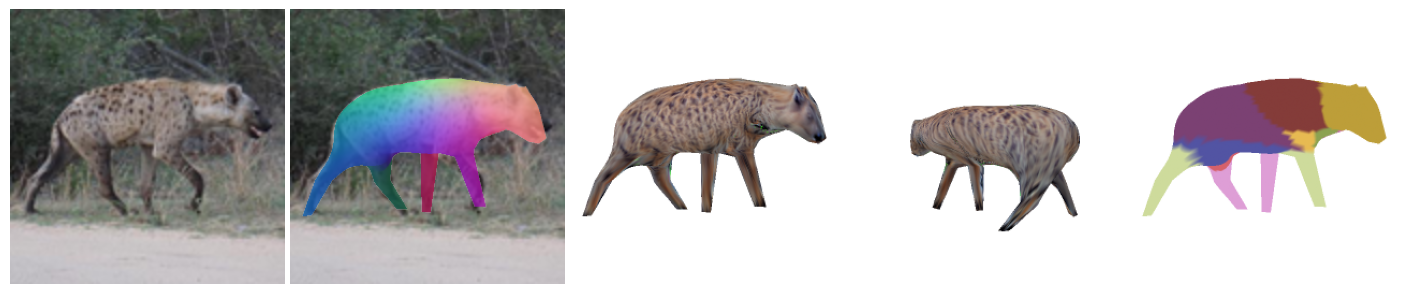}
    \includegraphics[width=0.4975\textwidth, trim=5 0 5 0, clip]{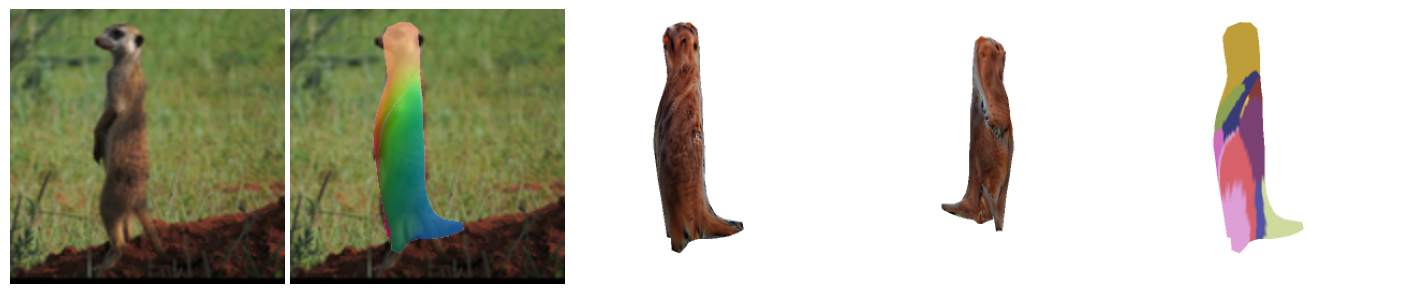}
    \vspace{-5pt}
    \includegraphics[width=0.4975\textwidth, trim=5 0 5 0, clip]{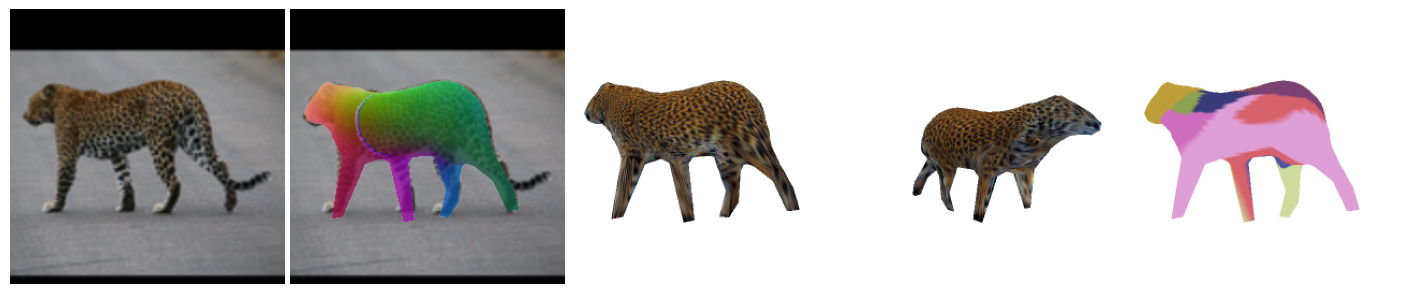}
    \includegraphics[width=0.4975\textwidth, trim=5 0 5 0, clip]{figs/sup_qual_saor_101/038.png}
    \vspace{-5pt}
    \includegraphics[width=0.4975\textwidth, trim=5 0 5 0, clip]{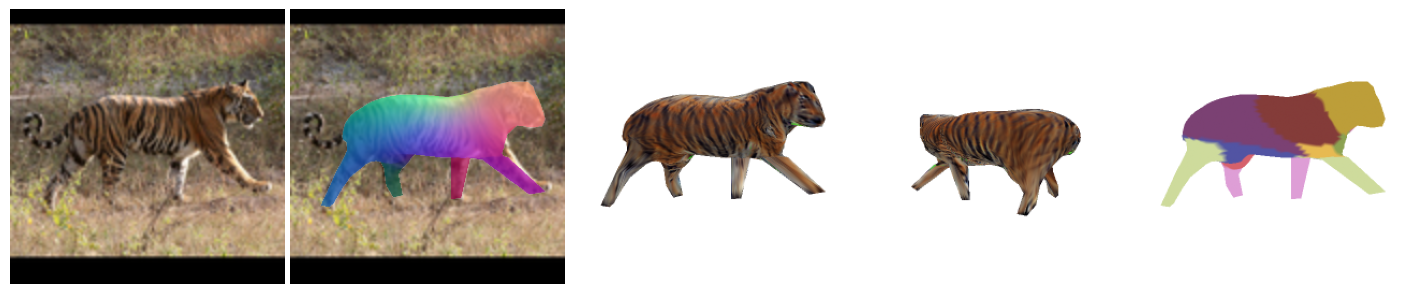}
    \includegraphics[width=0.4975\textwidth, trim=5 0 5 0, clip]{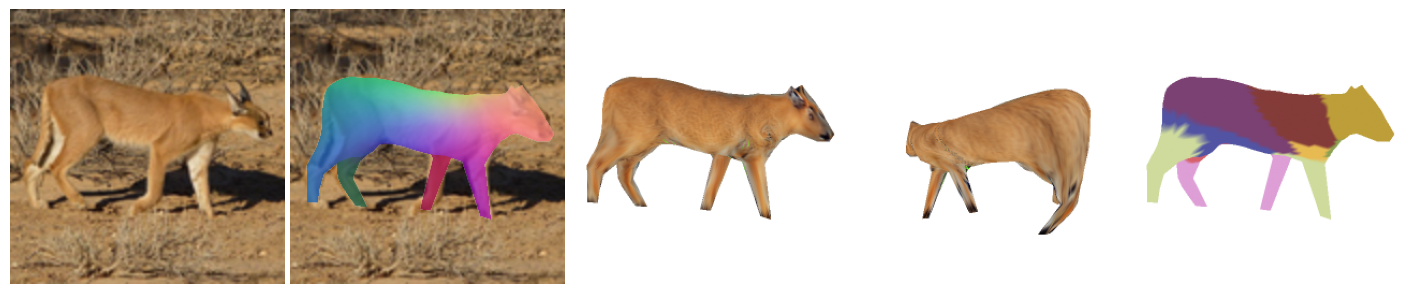}
    \vspace{-5pt}
    \includegraphics[width=0.4975\textwidth, trim=5 0 5 0, clip]{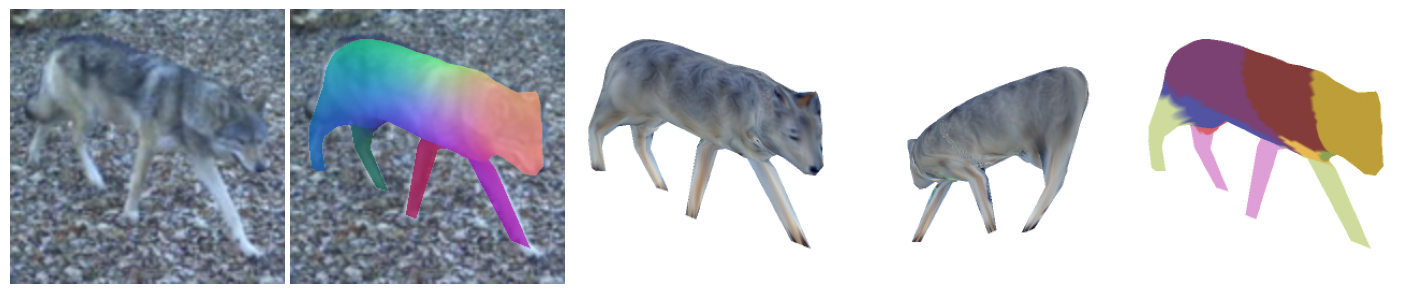}
    \includegraphics[width=0.4975\textwidth, trim=5 0 5 0, clip]{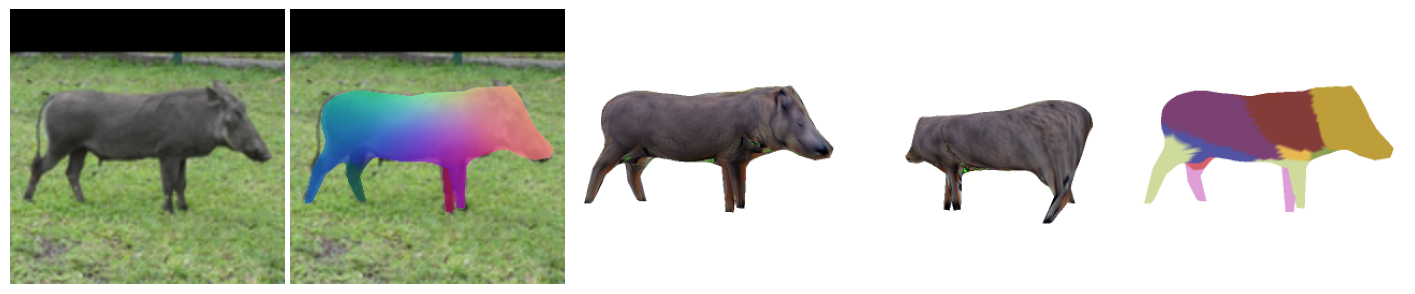}
    \vspace{-5pt}
    \includegraphics[width=0.4975\textwidth, trim=5 0 5 0, clip]{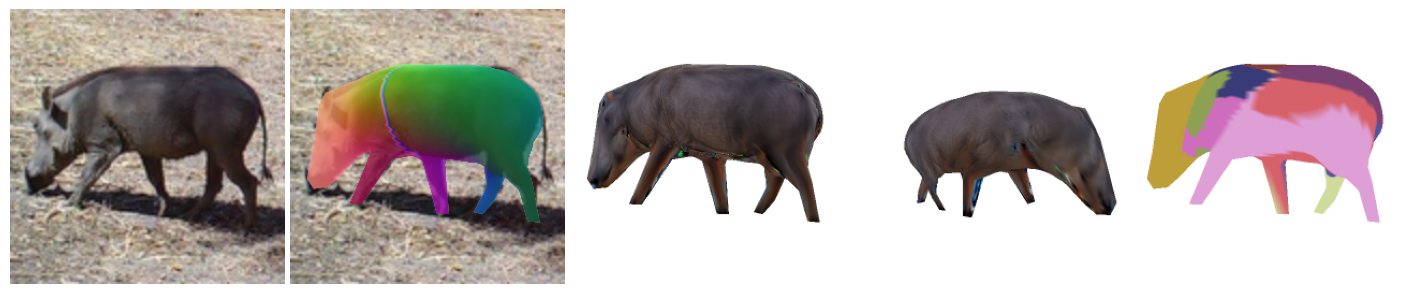}
    \includegraphics[width=0.4975\textwidth, trim=5 0 5 0, clip]{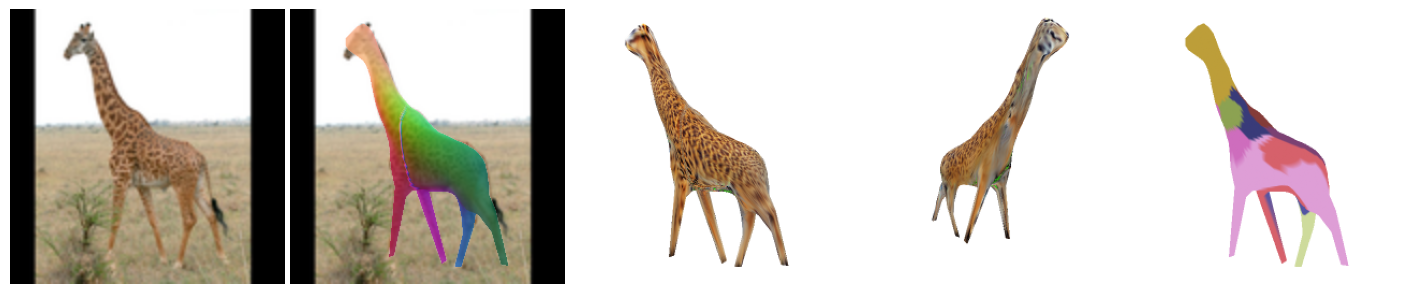}
    \vspace{-5pt}
    \includegraphics[width=0.4975\textwidth, trim=5 0 5 0, clip]{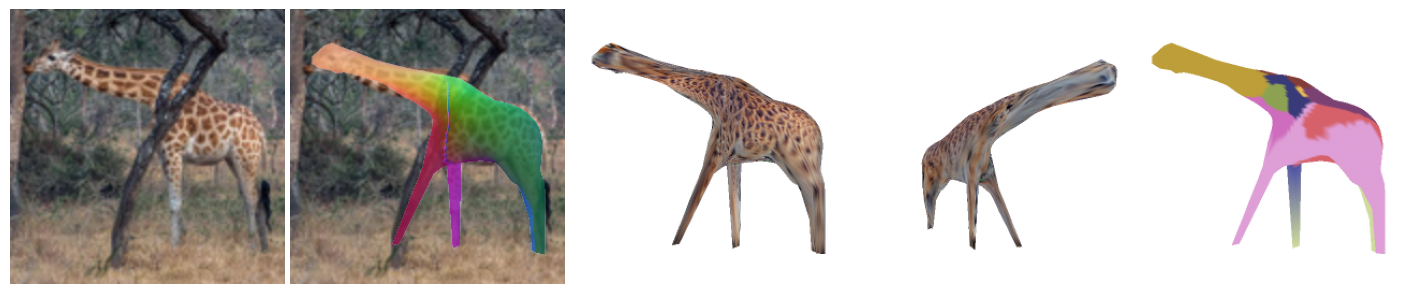}
    \includegraphics[width=0.4975\textwidth, trim=5 0 5 0, clip]{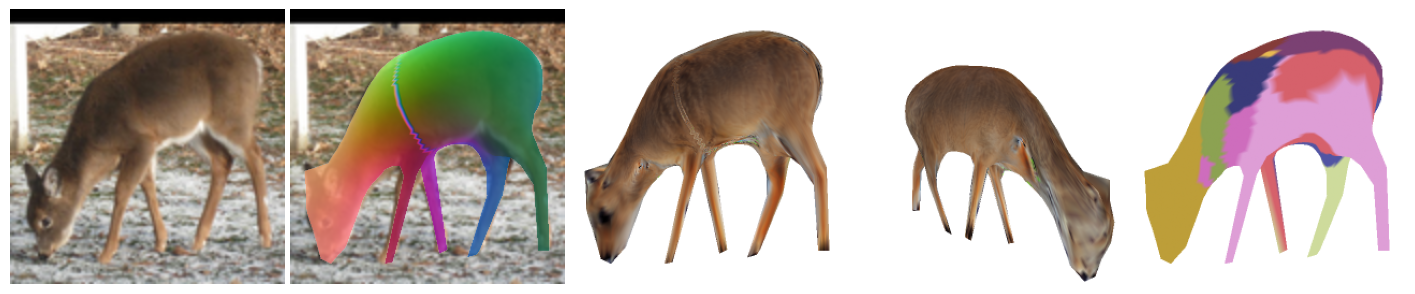}
    \vspace{-5pt}

\end{center} \vspace{-12pt}
\caption{Additional qualitative results for our SAOR approach on various different animal categories. 
Note  that the part assignment displays the part with the highest probability for each vertex, but in practice, the articulation for each vertex can be explained by a linear combination of multiple parts. }
\label{fig:sup_qual_saor_101_1}
\end{figure*}

\begin{figure*}
\begin{center}    

    \begin{picture}(0.49\textwidth, 5)
    \put(0.03\textwidth,0){Input}
    \put(0.12\textwidth,0){Pose}
    \put(0.24\textwidth,0){Reconstruction}
    \put(0.41\textwidth,0){Parts}
    \end{picture}
    \begin{picture}(0.49\textwidth, 5)
    \put(0.035\textwidth,0){Input}
    \put(0.13\textwidth,0){Pose}
    \put(0.24\textwidth,0){Reconstruction}
    \put(0.41\textwidth,0){Parts}
    \end{picture}
    \vspace{-5pt}
    \includegraphics[width=0.4975\textwidth, trim=5 0 5 0, clip]{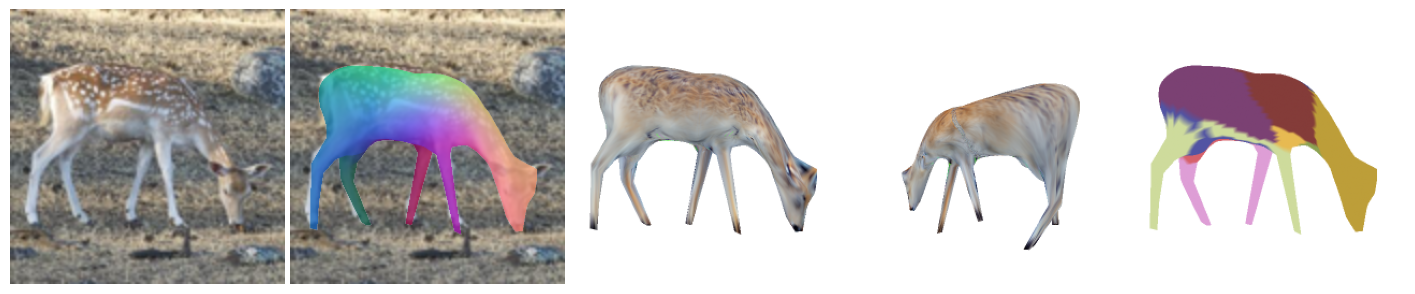}
    \includegraphics[width=0.4975\textwidth, trim=5 0 5 0, clip]{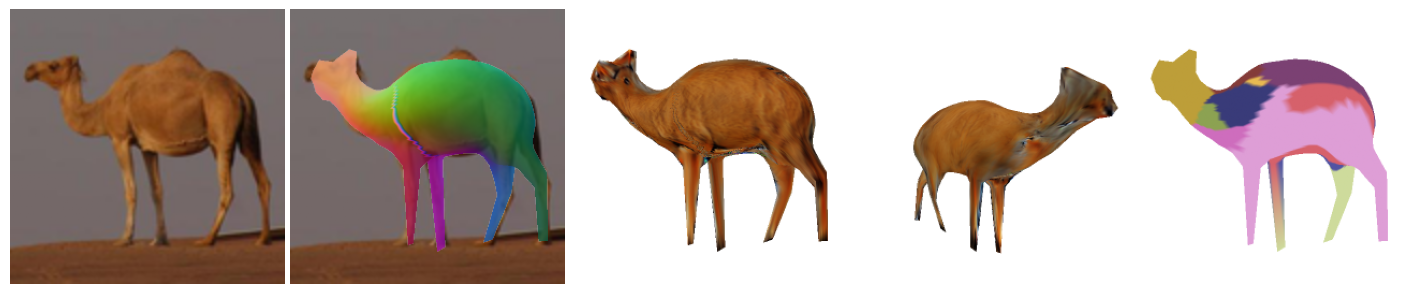}
    \vspace{-5pt}
    \includegraphics[width=0.4975\textwidth, trim=5 0 5 0, clip]{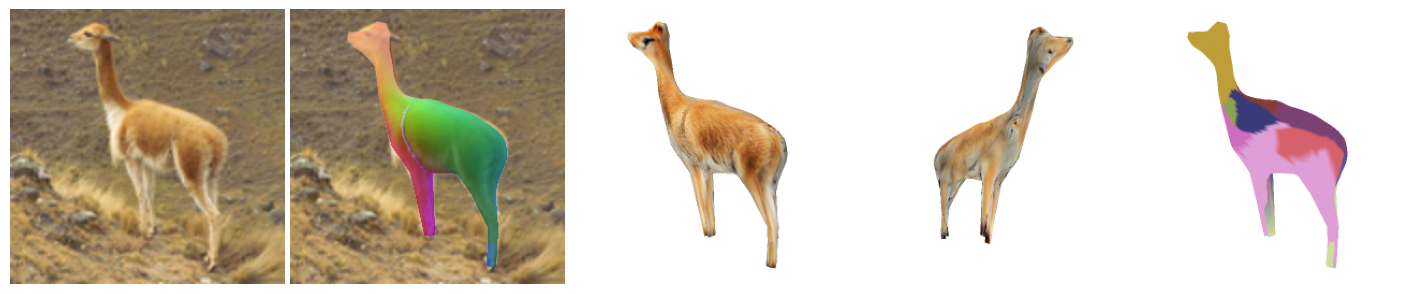}
    \includegraphics[width=0.4975\textwidth, trim=5 0 5 0, clip]{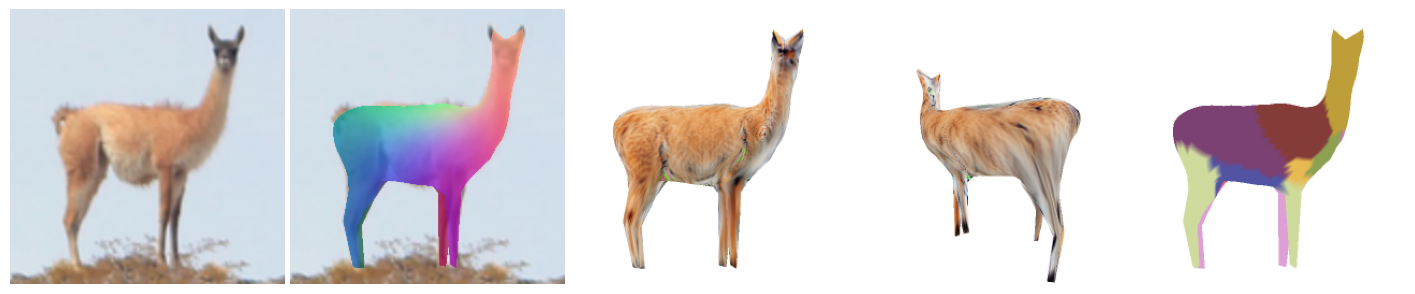}
    \vspace{-5pt}
    \includegraphics[width=0.4975\textwidth, trim=5 0 5 0, clip]{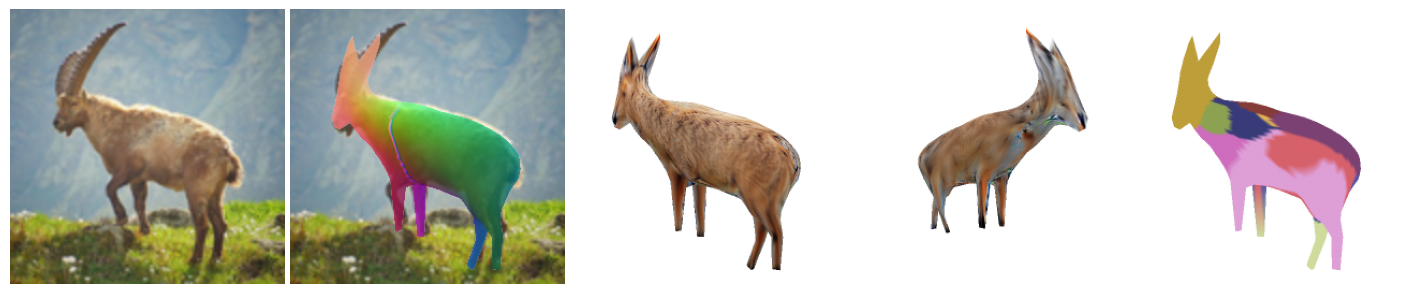}
    \includegraphics[width=0.4975\textwidth, trim=5 0 5 0, clip]{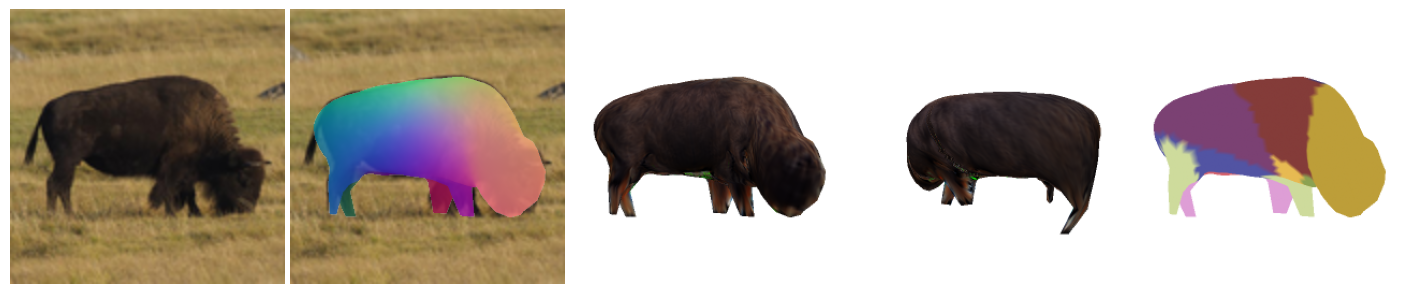}
    \vspace{-5pt}
    \includegraphics[width=0.4975\textwidth, trim=5 0 5 0, clip]{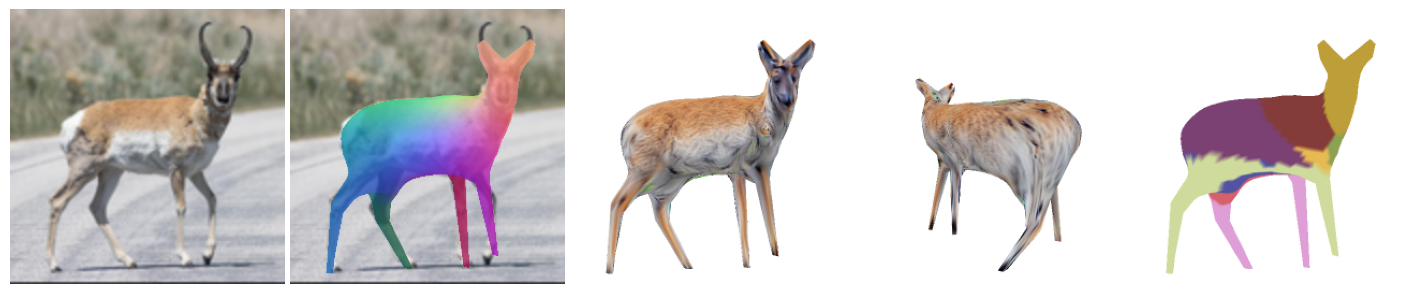}
    \includegraphics[width=0.4975\textwidth, trim=5 0 5 0, clip]{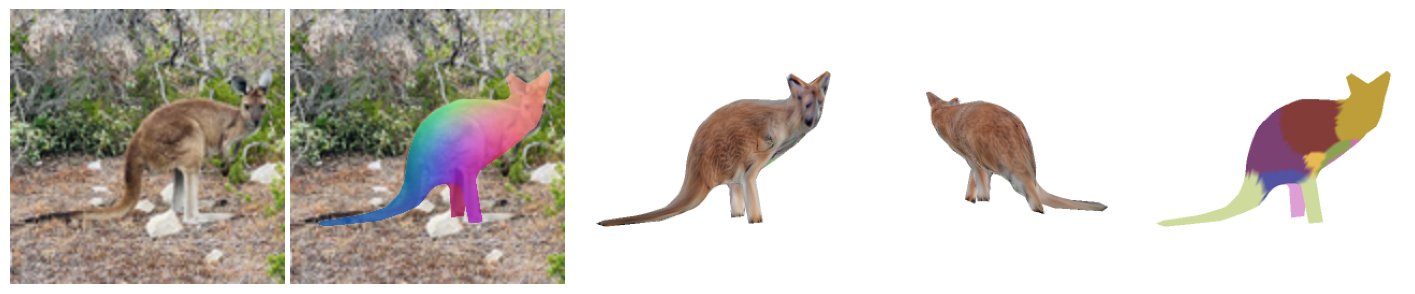}
    \vspace{-5pt}
    \includegraphics[width=0.4975\textwidth, trim=5 0 5 0, clip]{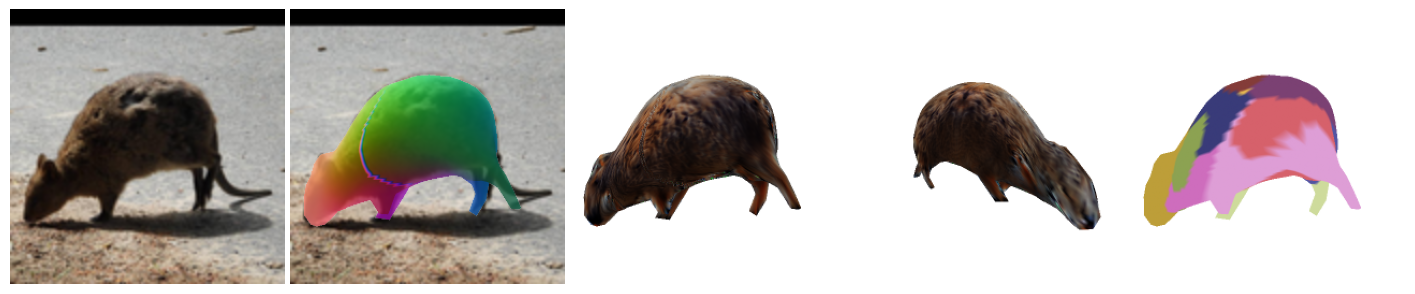}
    \includegraphics[width=0.4975\textwidth, trim=5 0 5 0, clip]{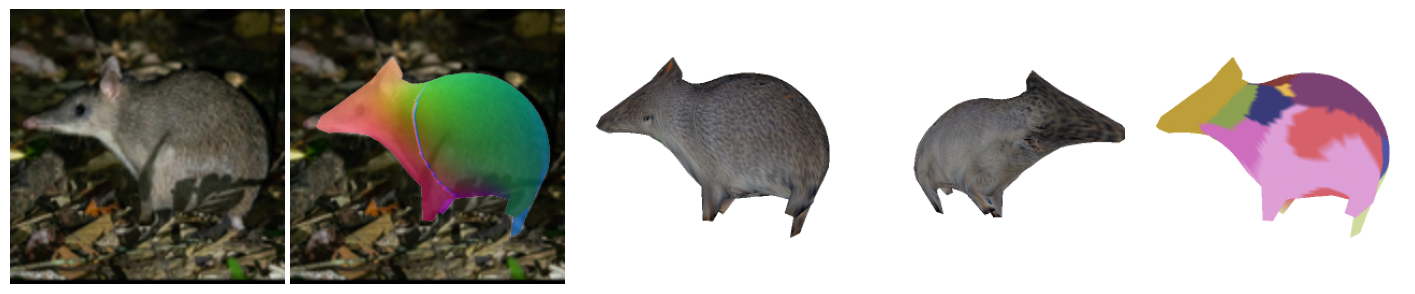}
    \vspace{-5pt}
    \includegraphics[width=0.4975\textwidth, trim=5 0 5 0, clip]{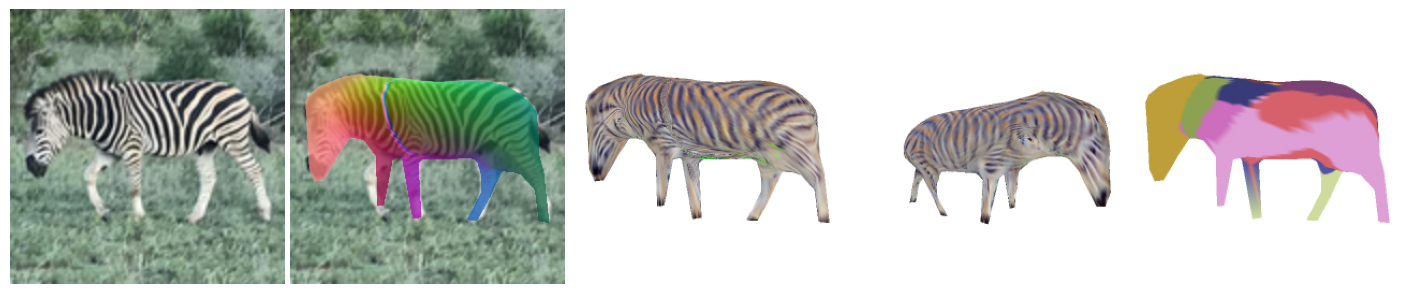}
    \includegraphics[width=0.4975\textwidth, trim=5 0 5 0, clip]{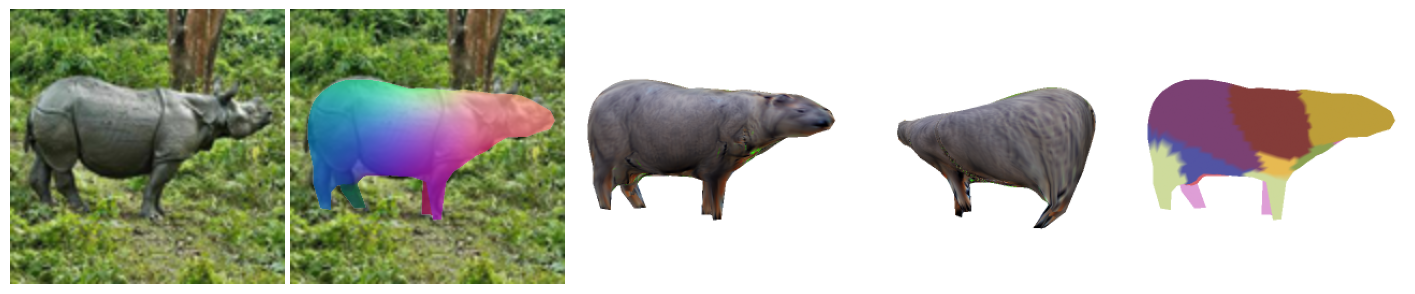}
    \vspace{-5pt}
    \includegraphics[width=0.4975\textwidth, trim=5 0 5 0, clip]{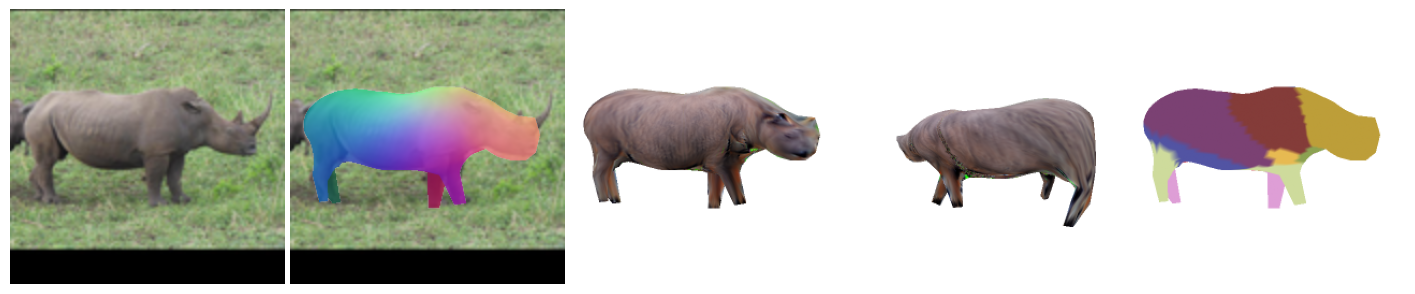}
    \includegraphics[width=0.4975\textwidth, trim=5 0 5 0, clip]{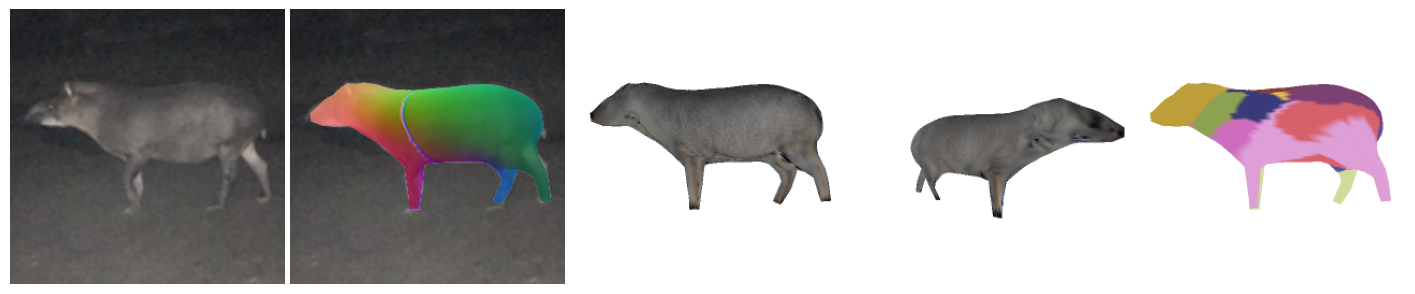}
    \vspace{-5pt}
    \includegraphics[width=0.4975\textwidth, trim=5 0 5 0, clip]{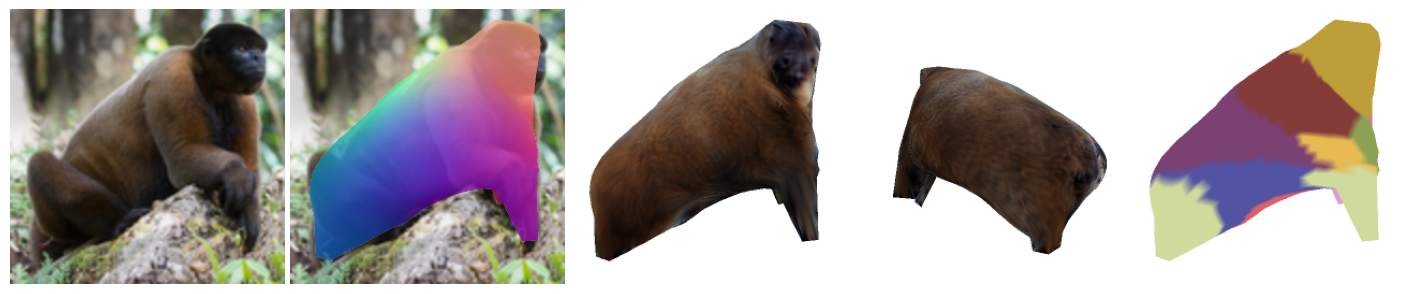}
    \includegraphics[width=0.4975\textwidth, trim=5 0 5 0, clip]{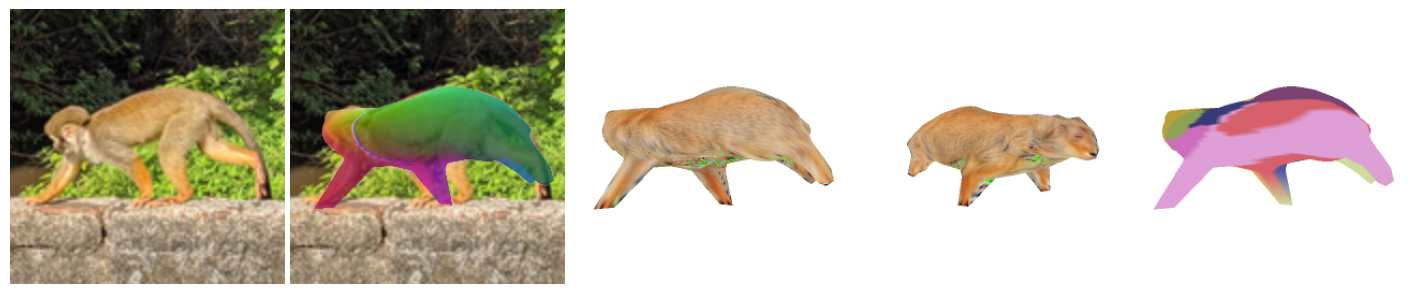}
    \vspace{-5pt}
    \includegraphics[width=0.4975\textwidth, trim=5 0 5 0, clip]{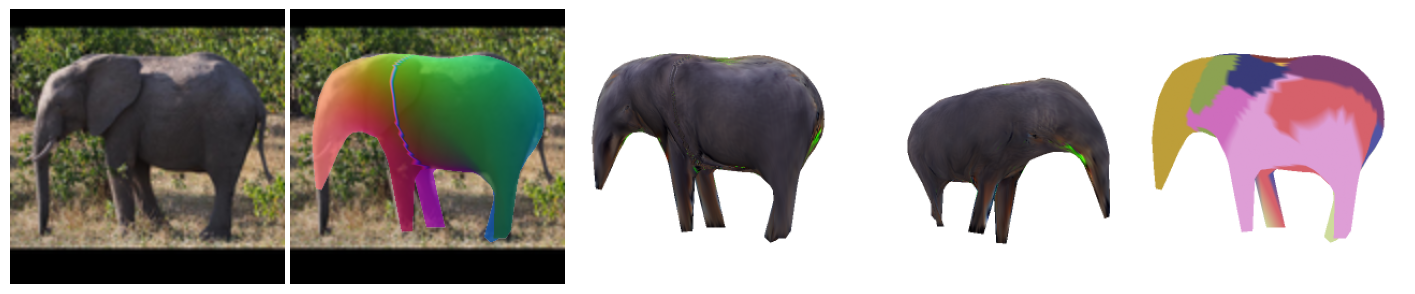}
    \includegraphics[width=0.4975\textwidth, trim=5 0 5 0, clip]{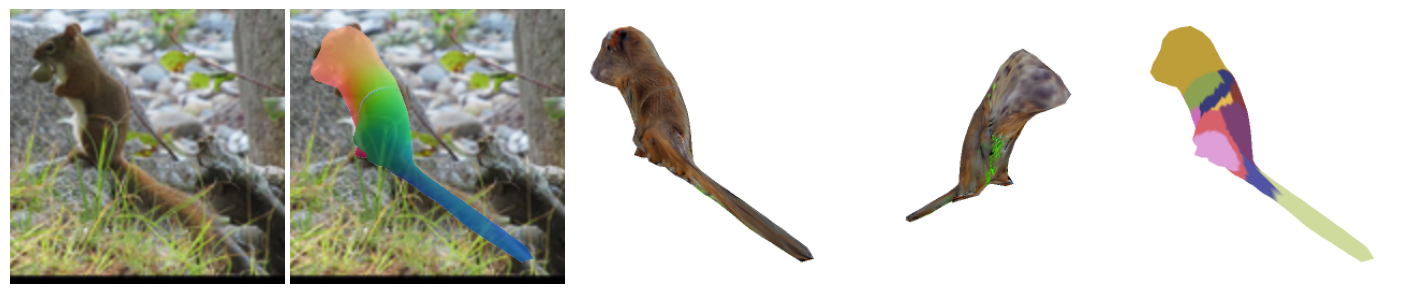}
    \vspace{-5pt}
    \includegraphics[width=0.4975\textwidth, trim=5 0 5 0, clip]{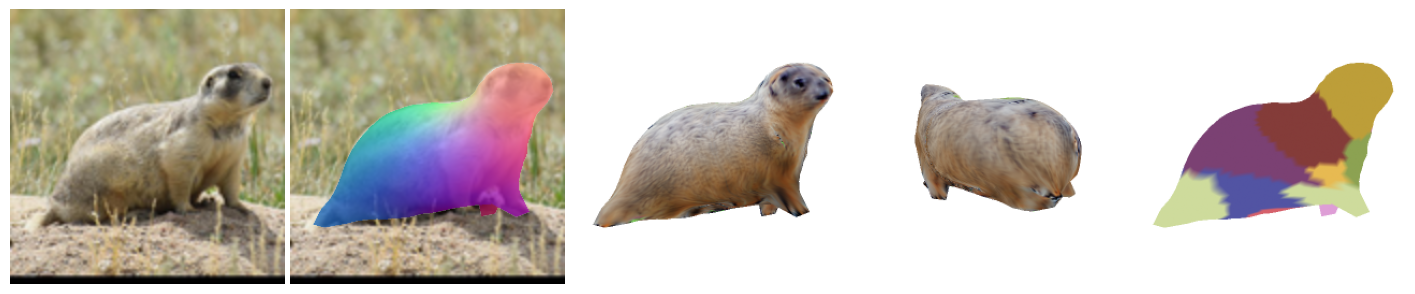}
    \includegraphics[width=0.4975\textwidth, trim=5 0 5 0, clip]{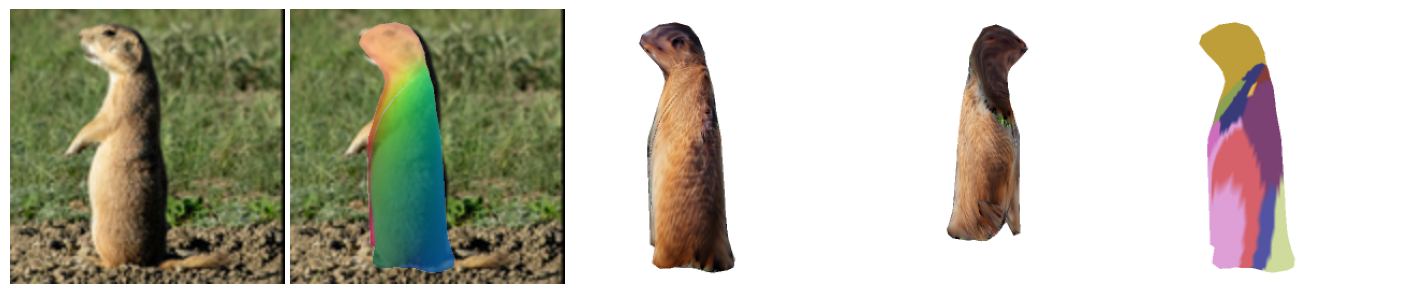}
    \vspace{-5pt}
    \includegraphics[width=0.4975\textwidth, trim=5 0 5 0, clip]{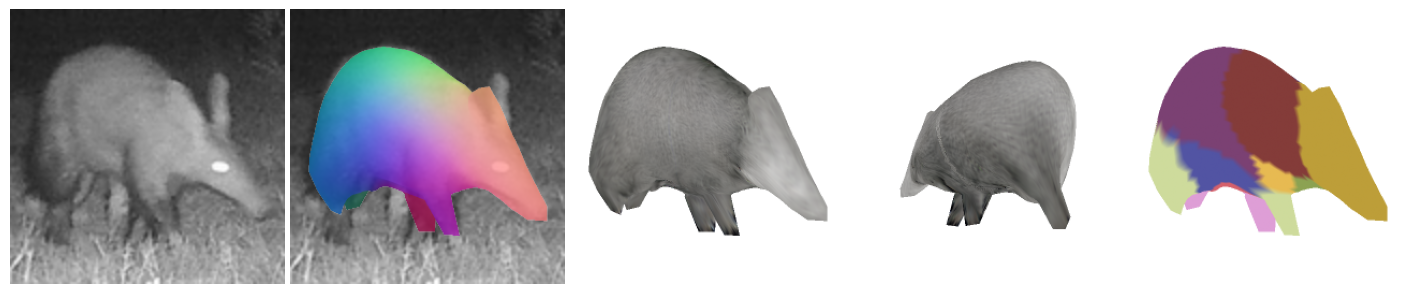}
    \includegraphics[width=0.4975\textwidth, trim=5 0 5 0, clip]{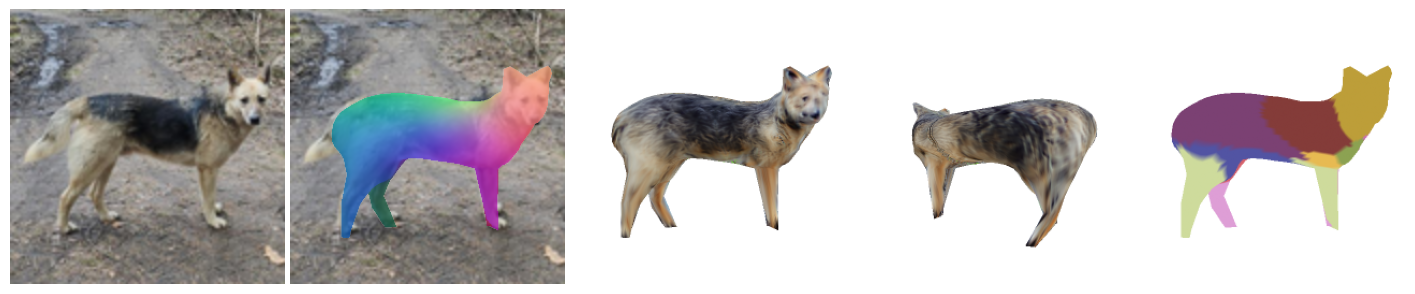}
    \vspace{-5pt}
    \includegraphics[width=0.4975\textwidth, trim=5 0 5 0, clip]{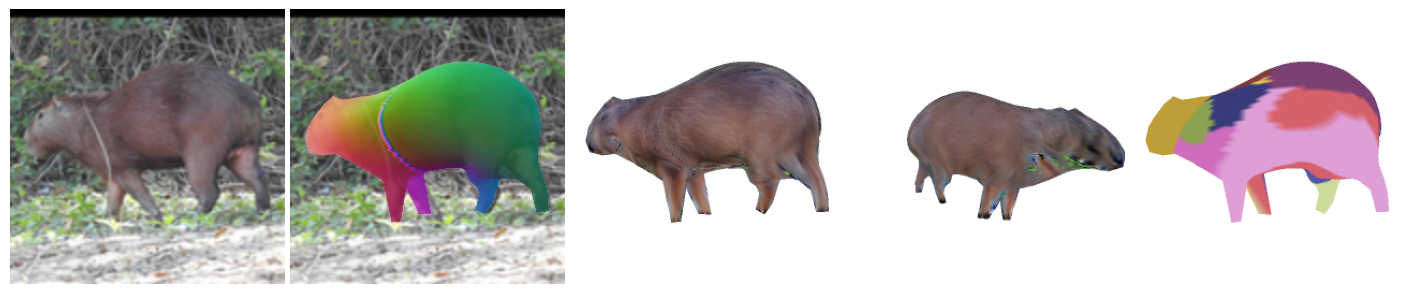}
    \includegraphics[width=0.4975\textwidth, trim=5 0 5 0, clip]{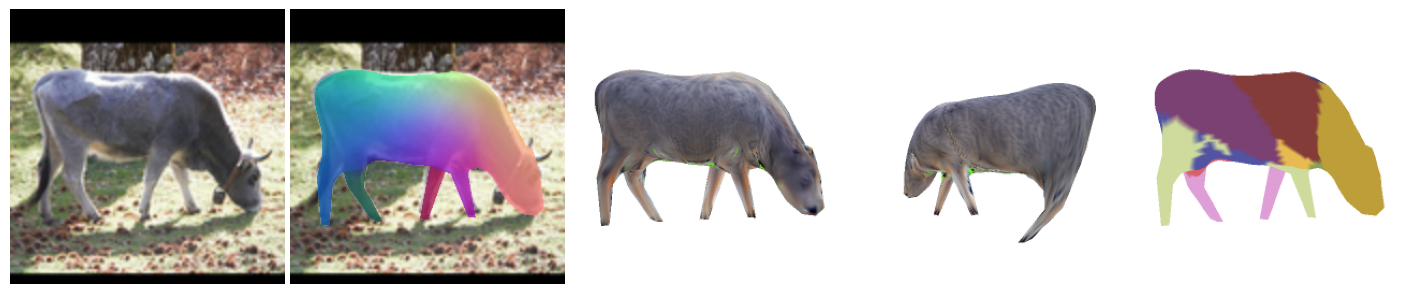}
    \vspace{-5pt}

\end{center} \vspace{-12pt}
\caption{Additional qualitative results for our SAOR approach on various different animal categories. 
Note  that the part assignment displays the part with the highest probability for each vertex, but in practice, the articulation for each vertex can be explained by a linear combination of multiple parts.}
\label{fig:sup_qual_saor_101_2}
\end{figure*}

\noindent{\bf Qualitative Results.} In Fig.~\ref{fig:sup_qual_saor_101_1} and Fig.~\ref{fig:sup_qual_saor_101_2} we present additional qualitative results on various animal categories all generate using our SAOR models that are trained on multiple categories. 
We provide additional results showing full 360-degree predictions for multiple different categories on the project website: \url{mehmetaygun.github.io/saor}. 

\begin{figure}
\begin{center}    
    \begin{picture}(\columnwidth, 10)
    \put(0.4\columnwidth,5){A-CSM~\cite{kulkarni2020acsm}}
    \end{picture}
    \vspace{8pt}
    \includegraphics[height=2.5cm, trim=0 0 0 0, clip]{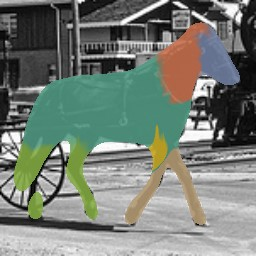}
    \includegraphics[height=2.5cm, trim=0 0 0 0, clip]{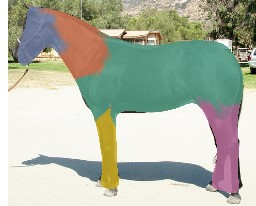}
    \includegraphics[height=2.5cm, trim=0 0 0 0, clip]{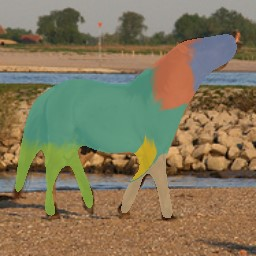}
    
    \begin{picture}(\columnwidth, 10)
    \put(0.40\columnwidth,5){SAOR-101}
    \end{picture}
    \includegraphics[height=2.5cm, trim=10 10 10 10, clip]{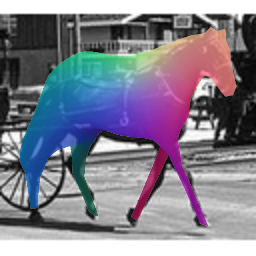}
    \includegraphics[height=2.4cm, trim=5 25 5 25, clip]{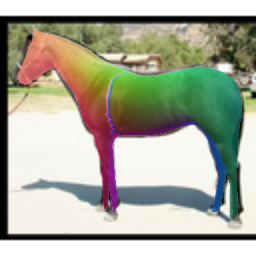}
    \includegraphics[height=2.5cm, trim=0 12 7 12, clip]{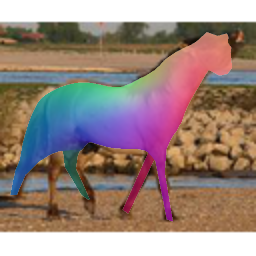}
    
\end{center} 
\vspace{-15pt}
\caption{Comparison with A-CSM~\cite{kulkarni2020acsm} on horses using example images from their paper. 
Even though A-CSM uses a 3D template with pre-defined fixed parts, it still maps left and right legs to the same leg in the template and the legs are not consistent across viewpoints (\ie the part assignment is different in the top row depending on whether the horse is facing left or right. 
In contrast, despite not using any 3D object priors at training time, our method is much more consistent in its assignment.  
However, it does mistake one of the left legs for the horse's tail in the final column. 
}
\label{fig:acsm}
\vspace{-10pt}
\end{figure}

\noindent{\bf Part Consistency.} We also compared SAOR's surface estimates with A-CSM~\cite{kulkarni2020acsm} in Fig.~\ref{fig:acsm}. 
Unlike A-CSM, our method does not use any 3D parts or 3D shape priors but is still able to capture finer details like discriminating left and right legs. 
A-CSM groups left and right legs as a single leg while their reference 3D template has left and right legs as a separate entity. 
Moreover, it mixes left-right consistency if the viewpoint changes. 

\begin{figure*}
\begin{center}    
    \begin{picture}(0.49\textwidth, 5)
    \put(0.03\textwidth,0){Input}
    \put(0.12\textwidth,0){Pose}
    \put(0.24\textwidth,0){Reconstruction}
    \put(0.41\textwidth,0){Parts}
    \end{picture}
    \begin{picture}(0.49\textwidth, 5)
    \put(0.035\textwidth,0){Input}
    \put(0.13\textwidth,0){Pose}
    \put(0.24\textwidth,0){Reconstruction}
    \put(0.41\textwidth,0){Parts}
    \end{picture}
    \vspace{-5pt}
    \includegraphics[width=0.4975\textwidth, trim=5 0 5 0, clip]{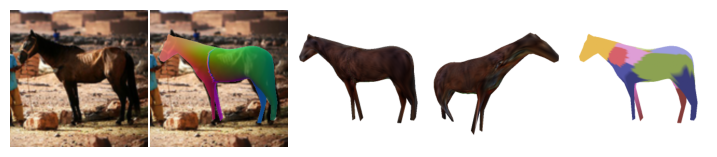}
    \includegraphics[width=0.4975\textwidth, trim=5 0 5 0, clip]{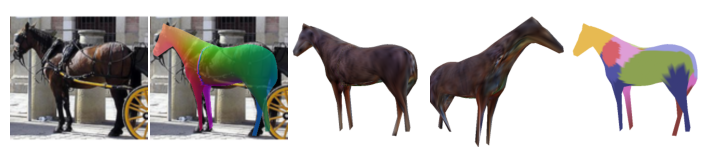}
    \begin{tikzpicture}
    \draw (0,0) -- (\textwidth,0);
    \end{tikzpicture}
    \vspace{-5pt}
    \includegraphics[width=0.4975\textwidth, trim=5 0 5 0, clip]{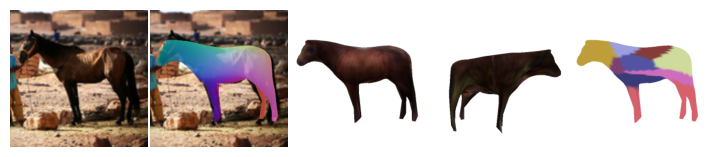}
    \includegraphics[width=0.4975\textwidth, trim=5 0 5 0, clip]{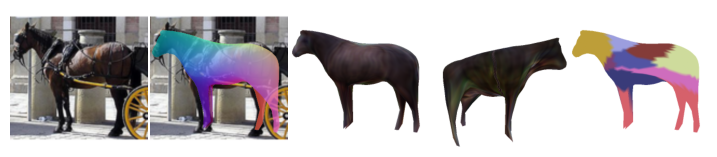}

\end{center} \vspace{-12pt}
\caption{Comparison of models trained with relative depth supervision (top) and without (bottom). Our model trained without depth also estimates detailed 3D shapes with the correct viewpoint. However, the 3D predictions are marginally worse as the model without depth produces slightly wider 3D shapes. Please note that part assignment and pose orientation are different since the two models started from different random initializations.}
\label{fig:depth_ablation}
\end{figure*}

\noindent{\bf Without Depth.} We also demonstrate examples from a variant of our model that was trained \emph{without} using relative depth map supervision in Fig.~\ref{fig:depth_ablation}. 
We observe that this model is still capable of estimating detailed 3D shapes with accurate viewpoints and similar textures as the full model. 
However, the model trained without depth maps tends to produce wider shapes compared to the full model.
Quantitative results for our model without relative depth are available in Table~2 in the main paper.

\begin{figure*}
\begin{center}    
    \begin{picture}(\textwidth, 10)
    \put(0.05\textwidth,2){Input}
    \put(0.15\textwidth,2){Pose}
    \put(0.30\textwidth,2){Reconstruction}
    \put(0.50\textwidth,2){Parts}
    \put(0.73\textwidth,2){MagicPony~\cite{wu2022magicpony}}
    \end{picture}
    
    \includegraphics[height=2.3cm, trim=5 0 5 0, clip]{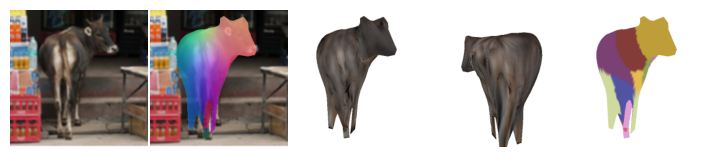}
    \includegraphics[height=2.3cm, trim=5 0 5 0, clip]{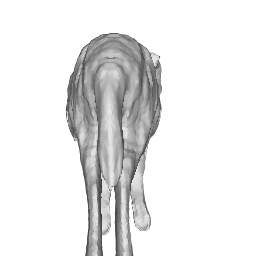}
    \includegraphics[width=2.3cm, trim=5 0 5 0, clip]{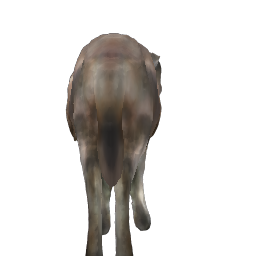}
    \includegraphics[width=2.3cm, trim=30 10 15 10, clip]{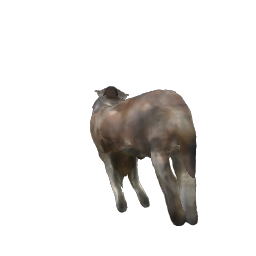}
    
    \includegraphics[height=2.3cm, trim=5 0 5 0, clip]{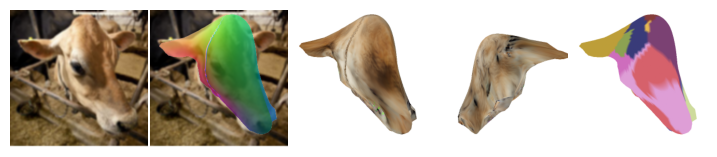}
    \includegraphics[height=2.3cm, trim=5 0 5 0, clip]{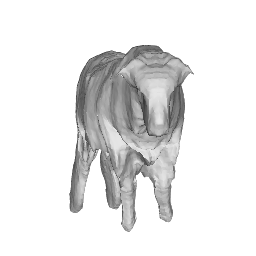}
    \includegraphics[width=2.3cm, trim=5 0 5 0, clip]{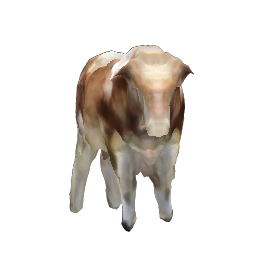}
    \includegraphics[width=2.3cm, trim=30 10 15 10, clip]{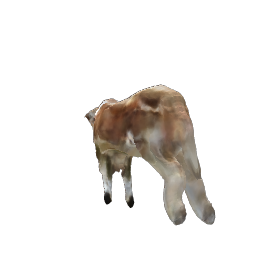}
 
    \vspace{-5pt}
   
\end{center} 
\vspace{-10pt}
\caption{Failure cases on cows. On the left we see SAOR-101 predictions (estimated pose, original viewpoint reconstruction, different view, and estimated parts). 
On the right we display MagicPony~\cite{wu2022magicpony} (original viewpoint reconstruction, textured reconstruction, different view). When the pose is very different than the typical ones present in the training set (top) or there is too much occlusion (bottom) our method fails to produce a sensible shape estimate. For the first example, MagicPony fails to capture the articulation of the head, and for the second occluded example it predicts an average template shape with the wrong pose.
}
\vspace{-5pt}
\label{fig:supp_fail}
\end{figure*}

\noindent{\bf Limitations.} We showcase some failure cases of our method in Fig.~\ref{fig:supp_fail}. 
Our method fails when the animal is captured from the back, as there is insufficient data available from that angle in the training sets. 
Note, methods such as~\cite{wu2022magicpony} partially address this by using alternative training data that includes image sequences from video. 
Furthermore, when there is also partial visibility (\eg only the head is visible), our method produces less meaningful results as our architecture does not explicitly model occlusion.

\noindent{\bf Part Ablations.} We conducted an additional ablation experiment on the number of parts used for horses. 
Results are provided in Table~\ref{tab:pck_part}. Notably, the PCK scores do not significantly vary with different numbers of parts. Therefore, for all other experiments, we used 12 parts.

\begin{table}[h]
    \centering

    \begin{tabular}{l c c c }
    \toprule
        Number of Parts & 6 & 12 & 24  \\ \midrule
        PCK & 43.8 & 44.9 & 44.1  \\ 
        \bottomrule
    \end{tabular}
    
    \vspace{-5pt}
    \caption{Keypoint transfer results on Pascal horses~\cite{everingham2015pascal} where the number of parts are varied.}
    \label{tab:pck_part}
    \vspace{-10pt}
\end{table}

\section{Additional Implementation Details}

\subsection{Data Pre-Processing}

When constructing our training datasets, we run a general-purpose animal  detector~\cite{beery2019efficient} and eliminate objects if any of the following criteria hold: i) the confidence of the detection is less than 0.8, ii) the minimum side of the bounding box is less than 32 pixels, iii) the maximum side of the bounding box is less than 128 pixels, and iv) there is no margin greater than 10 pixels on all sides of the bounding box. 

We then automatically extract segmentation masks using Segment Anything Model~\cite{kirillov2023segany} with the detected bounding box. 
We automatically estimate the relative monocular depth using the transformer-based Midas~\cite{Ranftl2021,Ranftl2022}, using their Large DPT model. 

\begin{figure*}
\begin{center}    
    \includegraphics[width=\textwidth, trim=5 5 5 5 , clip]{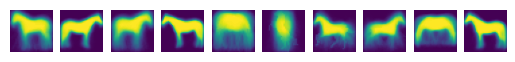}
    \includegraphics[width=\textwidth, trim=5 5 5 5 , clip]{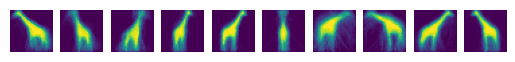}
    \includegraphics[width=\textwidth, trim=5 5 5 5 , clip]{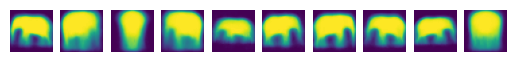}
     \includegraphics[width=\textwidth, trim=5 5 5 5 , clip]{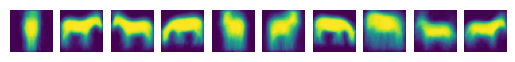}
    \includegraphics[width=\textwidth, trim=5 5 5 5 , clip]{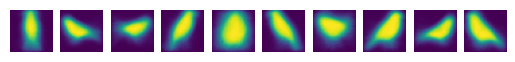}
\end{center} \vspace{-12pt}
\caption{Visualization of the cluster centers obtained from estimated silhouettes of various animal categories used in our balanced sampling. We observe that these cluster centers broadly capture the dominant viewpoints of each object category. Top to bottom: horse, giraffe, elephant, zebra, and bird.}
\label{fig:clusters}
\end{figure*}

To obtain cluster centers for the balanced sampling step in Section~3.3 in the main paper, we resize the estimated segmentation masks to $32\times32$, and cluster the 1024-dimensional vectors into 10 clusters using a Gaussian mixture model in all of our experiments. 
Visualization of cluster centers of various animals can be found in Fig.~\ref{fig:clusters}.

\subsection{Architecture}

We use a ResNet-50~\cite{he2016deep} as our image encoder $f_{enc}$ in our CUB\cite{CUB} experiments and the smaller ResNet-18 in quadruped animal experiments. 
This is in contrast to much larger ViT-based backbones used in other work~\cite{wu2022magicpony}. 
We initialize these encoders from scratch, \ie no supervised or self-supervised pre-training is used.  
The architecture details are presented in the following tables: deformation network $f_{d}$ in Table~\ref{tab:arch_def}, articulation network $f_{a}$ in Table~\ref{tab:arch_art}, texture network $f_{t}$ in Table~\ref{tab:arch_text}, and pose network $f_{p}$ in Table~\ref{tab:arch_pose}. 

\begin{table}
    \centering
    \begin{tabular}{l c c c}
    \toprule
          Layer & Input & Output & Dim  \\ \midrule
          Linear (3,512) & $S^{\circ}$  & $l_{x}$ &  $N\times512$\\
          Linear (512,512) & $\phi_{im}$  & $l_{z}$&  $1\times512$ \\
          2 $\times$ Linear (512,128) &  $l_{x} + l_{z}$  & $L$ &  $N\times128$\\
          Linear (128,3) &  $l$  & $D$ &  $N\times3$\\
          \bottomrule
    \end{tabular}
    \vspace{-5pt}
    \caption{Architecture details of our Deformation Net $f_{d}$. 
    } 
    \label{tab:arch_def}
\end{table}

\begin{table}
    \centering
    \begin{tabular}{l c c c}
    \toprule
          Layer & Input & Output & Dim  \\ \midrule
          Linear (3,512) & $S^{\circ}$  & $l_{x}$ &  $N\times512$\\
          Linear (512,512) & $\phi_{im}$  & $l_{z}$&  $1\times512$ \\
          Linear (512,128) &  $l_{x} + l_{z}$  & $L$ &  $N\times128$ \\
          Linear (128,128) &  $L$  & $L$ &  $N\times128$ \\
          Linear (128,K) &  $L$  & $W$ &  $N \times K$ \\
          K $\times$ Linear (512, 9) &  $\phi_{enc}$  & $\bf{\pi}$ &  $K \times 9$ \\
          \bottomrule
    \end{tabular}
    \vspace{-5pt}
    \caption{Architecture details of our Articulation Net $f_{a}$. $K$ is the number of parts and $N$ is the number of vertices, $\bf{\pi}$ is camera parameters. 
    }       
    \label{tab:arch_art}
\end{table}

\begin{table}
    \centering
    \begin{tabular}{l c c c}
    \toprule
          Layer & Input & Output & Dim  \\ \midrule
          Linear (512,512) & $\phi_{im}$  & $L$ &  $512\times1\times1$ \\
          Upsample  & $L$  & $L_{up}$ &  $512\times4\times4$ \\
          Upsample + Conv2D & $L_{up}$ & $L_{up}$  &  $256\times8\times8$ \\
          Upsample + Conv2D &  $L_{up}$ & $L_{up}$  &  $128\times16\times16$ \\
          Upsample + Conv2D &  $L_{up}$ & $L_{up}$  &  $64\times32\times32$ \\
          Upsample + Conv2D &  $L_{up}$ & $L_{up}$  &  $32\times64\times64$ \\
          Upsample + Conv2D &  $L_{up}$ & $L_{up}$  &  $16\times128\times128$ \\
          Conv2D & $L_{up}$ & T &  $3\times128\times128$ \\
          
          \bottomrule
    \end{tabular}
    \vspace{-5pt}
    \caption{Architecture details of our Texture Net $f_{t}$.} 
    \label{tab:arch_text}
\end{table}

\begin{table}
    \centering
    \begin{tabular}{l c c c}
    \toprule
          Layer & Input & Output & Dim  \\ \midrule
          1 $\times$ Linear (512,128) & $\phi_{im}$  & $L$ &  128\\
          C $\times$ Linear (128,6) & $L$  & $\bm{r}_{p},\bm{t}_{p}$ &  128\\
          Linear (128,C) & $L$  & $\bm{\alpha}$ &  128\\
          \bottomrule
    \end{tabular}
    \vspace{-5pt}
    \caption{Architecture details of our Pose Net $f_{p}$. C is the number of cameras, and $\bm{\alpha}$ are the associated scores for each camera~\cite{wu2022magicpony}.} 
    \label{tab:arch_pose}
\end{table}

\subsection{3D Evaluation Details}

For 3D quantitative evaluation, we used the Animal3D dataset~\cite{xu2023animal3d}. The dataset includes pairs of input images with their corresponding 3D models, which are estimated via optimizing the SMAL~\cite{zuffi20173d} model. Moreover, the 3D models are manually verified to eliminate poorly estimated shapes. 
We used the test split of the dataset for the horse, cow, and sheep categories. As there is no global pose alignment between our predictions and the dataset, we run the ICP algorithm to align them. We optimize rotation, $R\in\mathcal{R}^{3}$, translation $T\in\mathcal{R}^{3}$, and global scale $s\in\mathcal{R}^{1}$ with the Adam optimizer~\cite{kingma2014adam} using L1 norm as our alignment objective. We also follow the same alignment steps for the MagicPony~\cite{wu2022magicpony} baseline.

\subsection{Training Losses}
Here we describe the training losses from the main paper in more detail. 
The appearance loss is a combination of an RGB and perceptual loss~\cite{zhang2018unreasonable}.
$\mathcal{L}_{appr} = \lambda_{rgb}\mathcal{L}_{rgb} + \lambda_{percp}\mathcal{L}_{percp}$. These terms are defined below,

\begin{equation}
    \mathcal{L}_{rgb} = || \sum_{i,j} I_{i,j} - \hat{I}_{i,j} ||_{2}, 
\end{equation}

\begin{equation}
    \mathcal{L}_{percp} = || \phi_{p}(I_{i,j}) - \phi_{p}(I_{i,j}) ||_{2},
\end{equation}where $\phi_{p}$ is a function that extracts features from different layers of the VGG-16~\cite{simonyan2014very} network. 

The mask loss is calculated based on the difference between the automatically generated ground truth segmentation mask $M$ and the estimated mask $\hat{M}$ derived from our predicted 3D shape, 
\begin{equation}
    \mathcal{L}_{mask} = \lambda_{mask} \sum_{i,j} ||M_{i,j} - \hat{M}_{i,j}||_{2}.
\end{equation} 
Likewise, the depth loss is computed using the automatically generated relative depth $D$ and the estimated depth $\hat{D}$ from the predicted shape,
\begin{equation}
    \mathcal{L}_{depth} = \lambda_{depth}\sum_{i,j} ||D_{i,j} - \hat{D}_{i,j}||_{2}. 
\end{equation} 

Our swap loss is a combination of the RGB and mask loss between the input image $I$ and swapped image $I^{sw}$, 
\begin{equation}
    \mathcal{L}_{swap} = \lambda_{swap}\left[ \mathcal{L}_{mask}(I, I^{sw}) + \mathcal{L}_{rgb}(I, I^{sw})\right].
\end{equation} 

Finally, we also employ part regularization on the part assignment matrix $W$ to encourage equal-sized parts,
\begin{equation}
    \mathcal{L}_{part} = \lambda_{part}\sum_{k}^{K} \left((\sum_{i}^{N} W_{i,k}) - N/K\right)^2
\end{equation} 
where $N$ is the number of vertices in the mesh and $K$ is the number of parts. 
We also apply 3D regularization on the 3D shape, $\mathcal{L}_{smooth} = \lambda_{smooth}\sum_{}LS $, where $L$ is the laplacian of shape $S$ and $\mathcal{L}_{normal}$ which is defined below,

\begin{equation}
    \mathcal{L}_{normal} = \lambda_{normal} \sum_{\bf{n_{i}},\bf{n_{j}} \in \Omega} 
    1 - \frac{\bf{n_{i}}.\bf{n_{j}}}{||\bf{n_{i}}|| . ||\bf{n_{j}}||}
\end{equation} 
Here, $n_{i}$, $n_{j}$ are normals of neighbor faces. And the smoothness regularization is defined as $\lambda_{smooth}\mathcal{L}_{smooth} = ||LV||$, where $L$ is the Laplacian operator on the vertices. The final regularization term is defined as,
\begin{align}
    \mathcal{L}_{reg} = \lambda_{part} \mathcal{L}_{part} \newline 
    + \lambda_{smooth} \mathcal{L}_{smooth} + \lambda_{normal} \mathcal{L}_{normal}.
\end{align}
We note the weights used in our experiments for each loss in Table~\ref{tab:hyperparams}.

\subsection{Training}
 
In our experiments, we trained two different models: SAOR-101 and SAOR-Birds. The bird model is trained from scratch on CUB~\cite{CUB} for 500 epochs. In the first 100 epochs we only learn deformation, and then enable articulation afterwards.

The SAOR-101 model is trained in two steps. We first train the model using only Horse data from LSUN~\cite{yu2015lsun} then finetune it on all 101 animal categories downloaded from the iNaturalist website~\cite{iNatWeb}. In a similar fashion to the SAOR-Birds model,  we only learn deformation in the first 100 epochs, then allow articulation for about 300 epochs on horse data. Finally, fine-tune the model on all categories on iNaturalist data for 150 epochs. 
We utilize Adam~\cite{kingma2014adam} with a fixed learning rate for optimizing our networks. 
We note the hyperparameters used in Table~\ref{tab:hyperparams}.

Our simplified swap loss leads to easy hyper-parameter selection compared to Unicorn~\cite{monnier2022share}. 
For instance, in their swap loss term, the following parameters need to be decided: i) feature bank size, ii) minimum and maximum viewpoint difference, and iii) number of bins to divide samples in the feature bank depending on the viewpoint. 
Moreover, they need to do multistage training where they increase the latent dimensions for the shape and texture codes to obtain similar shapes during training. 
Here the number of stages and the dimension of latent codes in each stage are also hyperparameters. 
In our method, we eliminated all of these hyperparameters. 
Moreover, as we do not use all of the hypotheses cameras to estimate loss during a forward pass as in~\cite{wu2022magicpony} and as a result of our simplified swap loss, model training is six times faster than Unicorn, as they use six cameras during training, for the same number of epochs.

\begin{table}[h]
    \centering
      \resizebox{0.75\columnwidth}{!}{
    \begin{tabular}{l c}
    \toprule
          {\bf Parameter} & {\bf Value/Range}  \\ \midrule
          \textbf{Optimization} & \\ 
          Optimizer & Adam \\
          Learning Rate & 1e-4 \\ 
          Batch Size & 96 \\
          Epochs & 500 \\
          Image Size & 128 $\times$ 128 \\ \midrule
          \textbf{Mesh} & \\ 
          Number of Vertices & 2562 \\
          Number of Faces & 5120 \\
          UV Image Size & 64 $\times$ 128 $\times$ 3 \\ 
          Number of Parts & 12 \\
          Initial Position & (0,0,0) \\ \midrule
          \textbf{Camera} &  \\
          Translation Range & (-0.5, 0.5) \\
          Azim Range & (-180,180) \\
          Elev Range & (-15, 30) \\
          Roll Range & (-30, 30) \\ 
          FOV & 30 \\
          Number of Cameras & 4 \\ \midrule
          \textbf{Loss Weights} & \\
          $\lambda_{rgb}$ & 1 \\
          $\lambda_{percp}$ & 10 \\
          $\lambda_{mask}$ & 1 \\
          $\lambda_{depth}$ & 1 \\
          $\lambda_{swap}$ & 1 \\
          $\lambda_{smooth}$ & 0.1 \\
          $\lambda_{normal}$ & 0.1 \\
          $\lambda_{part}$ & 1 \\
          $\lambda_{pose}$ & 0.05 \\
          \bottomrule
    \end{tabular}
    }
    \vspace{-5pt}
    \caption{Training hyperparameters. }  
    \vspace{-15pt}
    \label{tab:hyperparams}
\end{table}

\end{document}